%% file: acl_latex.tex
\documentclass[11pt]{article}

\usepackage[final]{acl}

\usepackage{times}
\usepackage{latexsym}
\usepackage{amsmath}
\usepackage{xcolor}
\definecolor{CLEAR}{HTML}{7FC66E}
\usepackage{tikz}
\usepackage{multirow}
\usepackage{longtable}
\usepackage{multicol}
\definecolor{LAVENDER}{HTML}{AC89EC}
\definecolor{BLUE}{HTML}{47CAEE}
\definecolor{PINK}{HTML}{FF65F0}
\definecolor{CYAN}{HTML}{A5F4E2}
\definecolor{MultiToken}{HTML}{C2C2FF}

\usepackage{colortbl}
\usepackage{amssymb}

\usepackage[T1]{fontenc}
\usepackage[utf8]{inputenc}

\usepackage{microtype}
\newcommand{\circLabelBlue}[1]{%
\tikz[baseline=(char.base)]{
\node[shape=circle, draw=none, fill=BLUE,text=white, inner sep=1.5pt] (char) {#1};
}}

\newcommand{\circLabel}[1]{%
\tikz[baseline=(char.base)]{
\node[shape=circle, draw=none, fill=LAVENDER,text=white, inner sep=2pt] (char) {#1};
}}

\newcommand{\circLabelGreen}[1]{%
\tikz[baseline=(char.base)]{
\node[shape=circle, draw=none, fill=CLEAR,text=white, inner sep=2pt] (char) {#1};
}}

\newcommand{\circLabelPink}[1]{%
\tikz[baseline=(char.base)]{
\node[shape=circle, draw=none, fill=PINK,text=white, inner sep=2pt] (char) {#1};
}}

\newcommand{\circLabelCYAN}[1]{%
\tikz[baseline=(char.base)]{
\node[shape=circle, draw=none, fill=CYAN,text=black, inner sep=2pt] (char) {#1};
}}

\newcommand{\circLabelMULT}[1]{%
\tikz[baseline=(char.base)]{
\node[shape=circle, draw=none, fill=MultiToken,text=black, inner sep=2pt] (char) {#1};
}}
\usepackage{inconsolata}

\usepackage{graphicx}

\title{\textbf{\textcolor{black}{Dissecting Transformers:}} A \textbf{\textcolor{CLEAR}{CLEAR}} Perspective Towards Green AI}

\author{
  \textbf{Hemang Jain\textsuperscript{1*}} \quad
  \textbf{Shailender Goyal\textsuperscript{1*}} \quad
  \textbf{Divyansh Pandey\textsuperscript{1*}} \quad
  \textbf{Karthik Vaidhyanathan\textsuperscript{1}} \\
  \\
  \textsuperscript{1}IIIT Hyderabad \\
  \textsuperscript{*}Equal contribution
}

\vspace{-3cm}
\begin{document}
\maketitle
\begin{abstract}
The rapid adoption of Large Language Models (LLMs) has raised significant environmental concerns. Unlike the one-time cost of training, LLM inference occurs continuously and dominates the AI energy footprint. Yet most sustainability studies report only coarse model-level metrics, treating energy efficiency as an afterthought rather than a primary objective.
% We present the first fine-grained empirical analysis of inference energy across core components of Transformer Architecture. 
Addressing the limitation, we propose \textbf{Component-Level Energy Assessment via Repetitions (\textbf{\textcolor{CLEAR}{CLEAR}})}\footnote{Code: \href{https://anonymous.4open.science/r/CLEAR-D487}{https://anonymous.4open.science/r/CLEAR-D487}}, to overcome temporal mismatch between microsecond(\(\mu\) s) scale component execution and millisecond(ms) scale monitoring of energy sensors. Using \textbf{\textcolor{CLEAR}{CLEAR}}, we evaluate 15 models spanning four architecture types, keeping component-wise energy variance below \textbf{9.5\%} while capturing over \textbf{90\%} of total energy as individual components. 
% We present first fine-grained empirical analysis of different Transformer components across parameters like batch size, number of attention heads and hidden dimension size, impact of KV Cache and different variants and optimizations of Attention Mechanism. Our empirical analysis reveals that Attention blocks consume significantly more energy per FLOP, indicating that energy consumption is not proportionally aligned with FLOP counts. This shows that FLOPs alone fail to capture the true energy cost at a component level. \textbf{\textcolor{CLEAR}{CLEAR}} enables reliable energy measurements and provides a strong formal foundation for predictive modelling for energy consumption.
We present the first comprehensive, fine-grained energy analysis of Transformer components across key parameters such as batch size, attention heads, hidden dimension, KV cache, and attention variants. Our findings reveal that Attention consumes significantly more Energy per FLOP as compared to the entire model, indicating that FLOPs alone fail to capture true component-level energy cost. \textbf{\textcolor{CLEAR}{CLEAR}} enables reliable fine-grained energy measurements and provides a strong formal foundation for predictive modelling of energy consumption.
\end{abstract}

\section{Introduction}
\input{Files/Introduction}

\section{Related Works}
\input{Files/RelatedWorks}

\section{Methodology}
\input{Files/Methodology}

\section{Experimental Details}
\input{Files/ExperimentalDetails}

\section{Results}
\input{Files/Results}

\section{Discussion}
\input{Files/Discussion}

\input{Files/Ending}

\bibliography{custom}
\newpage
\vspace{2pt}
\newpage
\appendix

\input{Appendix/Models}
\input{Appendix/Transformers}

\input{Appendix/Hardware}

\input{Appendix/StdDev}

\input{Appendix/MultiToken}
\input{Appendix/Variants}
\input{Appendix/Tables/Tables_appendix}

\label{sec:appendix}

\end{document}

%% file: Files/Introduction.tex
Large language models (LLMs) \citep{openai2024gpt4technicalreport, grattafiori2024llama3herdmodels, yang2025qwen3technicalreport} have transformed natural language processing, but their environmental costs are increasingly concerning. ChatGPT-4o alone has been estimated to produce about 150,000 tons of CO$_2$e in 2025, equivalent to annual emissions of 30,000 gasoline powered cars or the carbon sequestration of a forest the size of city of Chicago ~\citep{jegham2025hungryaibenchmarkingenergy}.
While training and finetuning of LLMs is a computationally heavy and energy intensive process, it occurs infrequently. Inference, by contrast, happens continuously at a massive scale, with Gemini, GPT and Claude models serving hundreds of millions of queries daily . Consequently, even small reductions in per-inference energy consumption can translate into substantial global energy savings, making inference a critical target for optimizing the energy efficiency of LLMs. 

Current research focuses on model-level energy consumption with high-level comparisons across different systems~\citep{Alizadeh_2024,S_nchez_Momp__2025}. Recent works ~\citep{tian2025attentionsmicroscopecomparativestudy} increasingly rely on large datasets and report cumulative energy consumption across entire evaluation workloads rather than per-instance measurements.
Such aggregate reporting obscures the contribution of individual architectural components (e.g., Attention and MLP layers) to overall energy consumption. It also limits the understanding of how energy consumption scales with factors such as the number of tokens, batch size, hidden dimension size, and variations or optimizations of specific components. Fine-grained energy measurements are therefore necessary to enable targeted optimizations and support informed architectural design decisions.

In our work, we introduce \textbf{\textcolor{CLEAR}{CLEAR}}, a simple and adoptable three stage pipeline for fine-grained inference energy measurement of individual components in the
Transformer architecture. Unlike prior works limited to model and dataset level reporting ~\citep{faiz2024llmcarbonmodelingendtoendcarbon, jegham2025hungryaibenchmarkingenergy, Casta_o_2023}, \textbf{\textcolor{CLEAR}{CLEAR}}  decomposes the Transformer architecture into constituent components, such as Embedding layer, Normalization blocks, Attention, and feed-forward MLP and measures the energy consumed by each component. 
Our approach enables a systematic comparison of energy consumption patterns across model architectures, components, Floating Point Operations (FLOPs), hidden dimensions, batch sizes, input and output token lengths, as well as different variants and optimization strategies. 

\input{Figures/motivation}

A primary challenge arises from the temporal granularity mismatch between component execution and energy sensor monitoring as sub-operations complete within microseconds, whereas energy sensor provides power updates at tens of milliseconds. To address the gap, \textbf{\textcolor{CLEAR}{CLEAR}} employs an amplification strategy that stabilizes energy measurements. Another key challenge is the absence of a clear validation strategy for the reported energy measurements. \textbf{\textcolor{CLEAR}{CLEAR}} addresses the gap through a dedicated validation step based on two key metrics: (i) Completeness of the captured energy and (ii) Consistency across repeated trials. Our contributions are as follows :\\ 
\circLabelBlue{1} We propose a three stage pipeline, \textbf{\textcolor{CLEAR}{CLEAR}} (Component-Level Energy Assessment via Repetitions) to overcome temporal mismatch between microsecond ($\mu s$) scale component execution and millisecond (ms) scale energy sensors. \textbf{\textcolor{CLEAR}{CLEAR}} is deviced to measure energy consumption of fine-grained components in Transformer architecture. \\
\circLabelBlue{2} Using \textbf{\textcolor{CLEAR}{CLEAR}}, we empirically analyse 15 models spanning four key Transformer architectures, measuring energy consumption of fine-grained components including Attention, MLP, LM Head, and Normalization blocks. We further isolate the impact of key factors such as FLOPs, number of tokens, hidden dimension size, number of attention heads, and batch size on energy consumption. \\
\circLabelBlue{3} We conduct a controlled energy analysis to evaluate the impact of different Attention variants, optimization techniques, and KV Cache. Our findings reveal that energy consumed per FLOP varies substantially across model components, with the Attention block exhibiting the highest energy consumption per FLOP among the components analyzed.

%% file: Figures/motivation.tex
\begin{figure}[t]
    \centering
    \includegraphics[width=\columnwidth]{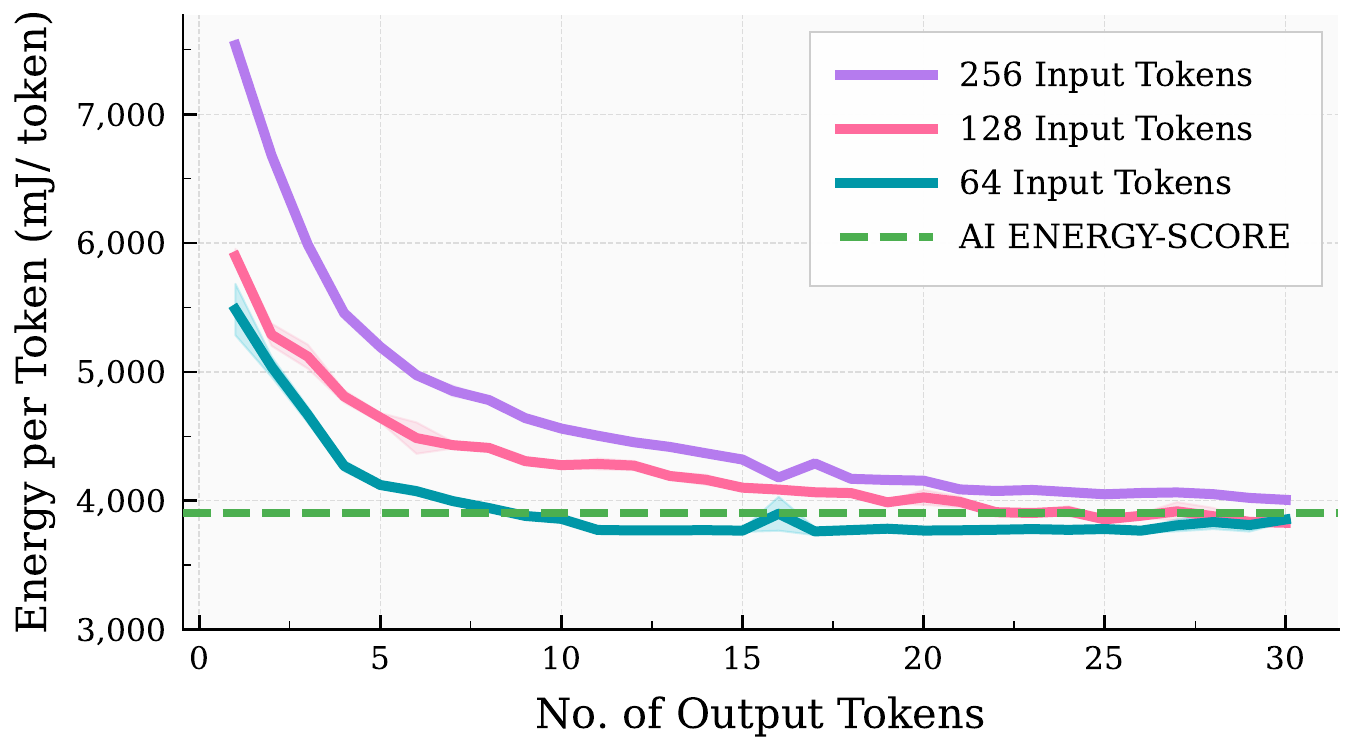}
    \caption{Comparison of the per-token energy reported by AI Energy Score~\cite{luccioni2025aienergyscore} with the fine-grained token-level measurements obtained using \textbf{\textcolor{CLEAR}{CLEAR}}.  Aggregate benchmarks obscure variability in fine-grained, token-level energy usage.}
    \label{fig:motivation}
    \vspace{-12pt}
\end{figure}

%% file: Files/RelatedWorks.tex
Prior works to measure energy consumption of computationally small components falls mainly in two categories: \circLabelPink{a} Methodologies that measure energy at a coarse granularity by measuring long sequences over large datasets and averaging to obtain per-token fine-grained approximations (Figure \ref{fig:motivation}) ~\citep{cao2021ireneinterpretableenergyprediction,schwartz2019greenai}. \citet{tian2025attentionsmicroscopecomparativestudy} reports the energy values over large datasets at the scale of MJ which fails to reliably isolate the true fine-grained effect for different Attention variants.
\circLabelPink{b} Hardware-centric approaches that rely on execution tracing are tightly coupled with specific sensors or system instrumentation \citep{jegham2025hungryaibenchmarkingenergy,vandervlugt2025powersensor3fastaccurateopen,ALVI2021100594,10.1145/2962131}. The additional sensor used needs to have low latency and high precision to reliably trace the function execution. Additionally, hardware based approaches are not scalable as they depend on specialized equipment and physical access to every target device making them difficult to deploy at scale. \textbf{\textcolor{CLEAR}{CLEAR}} departs from both paradigms by providing a component-level, software-based method validated for completeness and consistency. Operating entirely at application layer, \textbf{\textcolor{CLEAR}{CLEAR}} can be readily adopted across different hardware configurations and enables broader experimental space for energy analysis. 

To improve accessibility, lightweight monitoring tools such as CodeCarbon and Carbontracker~\cite{anthony2020carbontrackertrackingpredictingcarbon} have emerged but, they still operate at a coarse granularity (\circLabelPink{a}). More fine-grained approaches like  \textsc{FECoM}~\cite{rajput2024enhancingenergyawarenessdeeplearning} and \textsc{EdgeProfiler}~\cite{pinnock2025edgeprofilerfastprofilingframework},are targetted towards Tensorflow APIs and edge deployments respectively.  

Recent works like \citet{ozcan2025quantifyingenergyconsumptioncarbon} uses GPU-based simulations to study how batch size, sequence length, and parallelism influence inference efficiency. Extending real workloads, ~\citet{fernandez2025energyconsiderationslargelanguage} distinguish prefill and autoregressive stages, showing that optimizations can reduce energy consumption by up to 73\%. Broader benchmarks, like \emph{How Hungry is AI?}~\cite{jegham2025hungryaibenchmarkingenergy}, evaluate energy, water, and CO$_2$ footprints across hardware platforms, while the BLOOM case study. ~\citet{JMLR:v24:23-0069} was among the first to track emissions during the training and inference of a 176B parameter model. ~\citet{luccioni2025aienergyscore} benchmarks over 160 models across multiple tasks, reporting GPU energy consumption in deep learning models. However, unlike prior works that report aggregate model-level energy consumption, \textbf{\textcolor{CLEAR}{CLEAR}} adopts a granular perspective, decomposing the Transformer architecture into its constituent components to analyze energy consumption across key axes.

%% file: Files/Methodology.tex
\textbf{\textcolor{CLEAR}{CLEAR}} uses a simple and easily adoptable, three-stage pipeline comprising the following stages: \circLabelGreen{1} \textbf{Activation Store} to capture the sample activations, \circLabelGreen{2} \textbf{Amplification Strategy} for reliable energy measurement and \circLabelGreen{3} \textbf{Validation} to verify the Completeness and Consistency of the energy analysis. \textbf{\textcolor{CLEAR}{CLEAR}} targets key computational primitives common to most transformer-based models. These include the \textbf{Attention} block that captures token-level dependencies, feed-forward \textbf{MLP} blocks for dense nonlinear transformations, \textbf{Normalization blocks} (Norm.), the \textbf{Embedding Layer} which maps discrete tokens into continuous vector spaces and the final \textbf{Language Modeling Head} (LM Head) that projects hidden representations back to the vocabulary space for output generation. 

\subsection*{\circLabelGreen{1} Activation Store}
The Activation Store $\mathbf{A}$ serves as a cache of activations that allows isolated re-execution of individual components under identical input statistics, enabling fine-grained measurement of energy consumption. To enable component-wise energy profiling, we insert forward hooks at key points of the computation graph, 
\[
\mathcal{A} = \{\texttt{attn}, \texttt{mlp},
\texttt{lm\_head}, \texttt{layer\_norm} \ldots\}
\]
and capture the input activations at each hook.  
During a single forward pass, the hooks record for every component $c \in \mathcal{A}$ the corresponding activation tensor $a_c$ for all tokens,
\[
\mathbf{A} = \{a_c \mid c \in \mathcal{A}\}, \qquad 
a_c  \in \mathbb{R}^d .
\]

\input{Figures/methodology}

\subsection*{\circLabelGreen{2} Amplification Strategy}
Accurate energy measurement of Transformer components is challenging since the individual execution time is typically much shorter ($\mu s$) than the sampling period of GPU power sensors. For example, NVIDIA's NVML has a sensor read rate of about 20 to 50\,ms.  
The temporal mismatch leads to two distinct sources of error (Refer Figure \ref{fig:methodology}):

In case \circLabelCYAN{a}, when the component completes execution within a few microseconds, entirely between two sensor samples, the monitoring sensor cannot update its reading in time due to which the observed energy is reported as zero:
\begin{equation}
    E_2 - E_1 = 0,
\end{equation}
even though the true component energy is non-zero. This leads to the energy consumption of smaller components being neglected and unaccounted.

In case \circLabelCYAN{b}, if we supposedly measure energy after every sensor reading to capture the component’s energy consumption, the result remains highly noisy. This is because the measurement inevitably includes a significant amount of idle energy drawn by CUDA, making it hard to separate the true component energy. Consequently, when the execution only partially overlaps with a sensor’s sampling window, the observed energy is recorded as
\begin{equation}
    E_2 - E_1 = E_c \pm \varepsilon,
\end{equation}
where $E_c$ is the component's actual energy consumed and $\varepsilon$ represents noise. 

\input{Figures/sampling}

To address above challenges, we adopt an amplification strategy, illustrated in \circLabelCYAN{c} of Figure~\ref{fig:methodology}.  
As individual transformer components often complete the execution within $10$--$100\,\mu s$, their energy consumption remains highly noisy to NVML’s coarse sampling window. 
% The goal is to minimize noise ($\varepsilon$) in the component energy measurement to obtain reliable readings. 
To obtain reliable readings, it is essential to minimize noise ($\varepsilon$) in the component energy measurement. This noise may arise from the model’s idle energy consumption or from inherent errors in the sensor measurements. 
To achieve this, each component is executed repeatedly in rapid succession on cached activations, without gaps between runs. This approach scales the effective runtime so that the total energy of the repeated executions dominates the idle background consumption, rendering the noise comparatively negligible.

% which scales execution time so that the signal dominates noise. To overcome this, we rerun each component repeatedly on cached activations. Concretely, each component is executed $N = 10{,}000$ times within a single measurement window:
% \begin{equation}
%     E^{\text{tot}}_c = \text{MeasureEnergy}\!\left(\sum_{i=1}^N c(a_c)\right),
% \end{equation}
% where $a_c$ denotes the cached input activations and $c(\cdot)$ represents the component function. The per-execution energy is then obtained as:
% \begin{equation}
%     \hat{E}_c = \frac{E^{\text{tot}}_c}{N} \pm \frac{\varepsilon}{N}.
% \end{equation}

% This amplification ensures that the aggregated workload duration spans atleats hundreds of milliseconds, long enough to be captured by NVML’s sampling resolution. As per the above formula, if we keep increasing the the number of repetions N, the noise component goes on to decrease causing comparatively reliable measurements. 

Concretely, for each component $c$ with cached input $a_c$, we measure the energy before and after $N$ consecutive executions:
\begin{equation}
   E^{\text{tot}}_c = \mathrm{MeasureEnergy}\!\left(\sum_{i=1}^{N} c(a_c)\right).
\end{equation}
The per-execution energy can then be obtained by averaging the total measured energy:
\begin{equation}
   \hat{E}_c
   = \frac{E_{\text{end}} - E_{\text{start}}}{N}
   = \frac{E^{\text{tot}}_c}{N}
   \pm \frac{\varepsilon}{N},
\end{equation}
where $\varepsilon$ denotes the measurement noise. By increasing $N$, the duration of the aggregated workload extends to hundreds of milliseconds such that execution time is larger than reading frequency of  NVML’s power sensor while the noise term $\varepsilon/N$ diminishes proportionally, yielding significantly reliable per-component energy measurements \citep{Arafa_2020} as shown in \circLabelCYAN{c}. We repeat the amplified measurement for $T$ trials with a brief pause between runs to let the sensor reset, taking Average and Standard Deviation across trials to further smooth out sensor noise and make the per-component energy measurement more reliable. 
% (Refer Algorithm \ref{alg:clear})

% To average and smoothen out noise we further repeat the whole measurement $T$ times (with brief sleeps to reset the sensor) and report the mean and standard deviation,
% \[
%    \overline{E}_c=\tfrac{1}{T}\sum_{t=1}^T \hat{E}_c^{(t)},\qquad
%    \operatorname{Std}(\hat{E}_c)=\sqrt{\tfrac{1}{T}\sum_{t=1}^T(\hat{E}_c^{(t)}-\overline{E}_c)^2}.
% \]
% These two levels of averaging suppress noise and yield a more reliable per-component energy estimate.

\subsection*{\circLabelGreen{3} Validation}
Due to the lack of fine-grained energy analyses of individual components within Transformer architectures, the validation step in \textbf{\textcolor{CLEAR}{CLEAR}} verifies the reliability of the measured energy along two key axes: (1) Consistency across repeated trials and (2) Completeness of the captured energy.
\[
    \mathrm{StdDev}\bigl(\bar{E}_{\text{c}}\bigr) \to 0,  \qquad
    \bar{E}_{\text{model}} \approx \sum_{c \in \mathcal{C}} \bar{E}_c .
\]
 \circLabel{i} A standard deviation close to 0 indicates that repeated component-level energy measurements remain consistent across trials demonstrating high precision in energy measurements by CLEAR. 
 \circLabel{ii} The near-equality between the total measured model energy and sum of its per-component energies
demonstrates that CLEAR is able to capture the component’s energy usage in a comprehensive manner.

%% file: Figures/methodology.tex
\begin{figure}[t]
    \centering
    \includegraphics[width=\columnwidth]{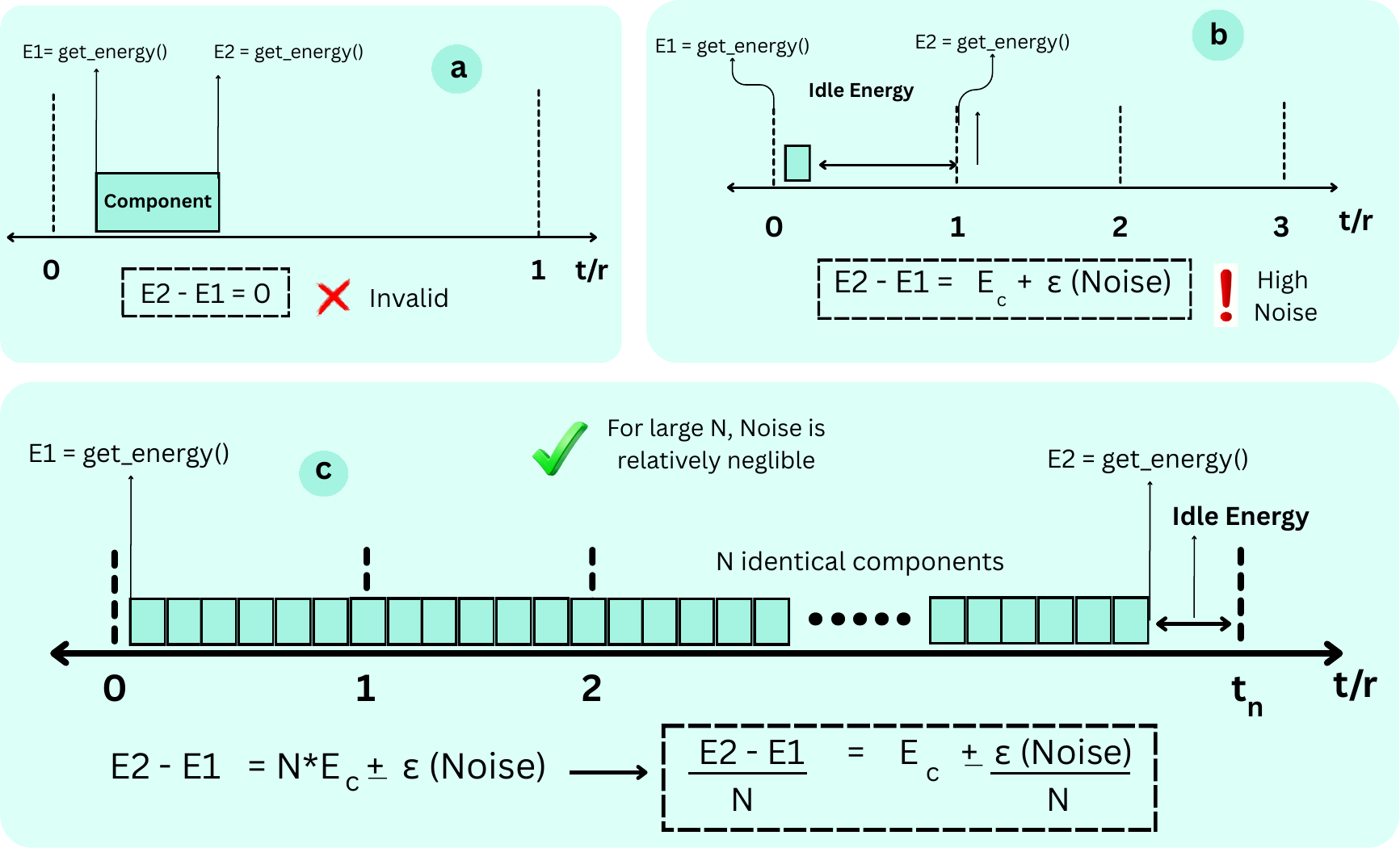}
    \caption{Comparison of Energy Sampling approaches. Case (c) demonstrates the sampling strategy followed by \textbf{\textcolor{CLEAR}{CLEAR}} to reduce the noise}
    \label{fig:methodology}
    \vspace{-10pt}
\end{figure}

%% file: Figures/sampling.tex
\begin{figure}[t]
    \centering
    \includegraphics[width=\columnwidth]{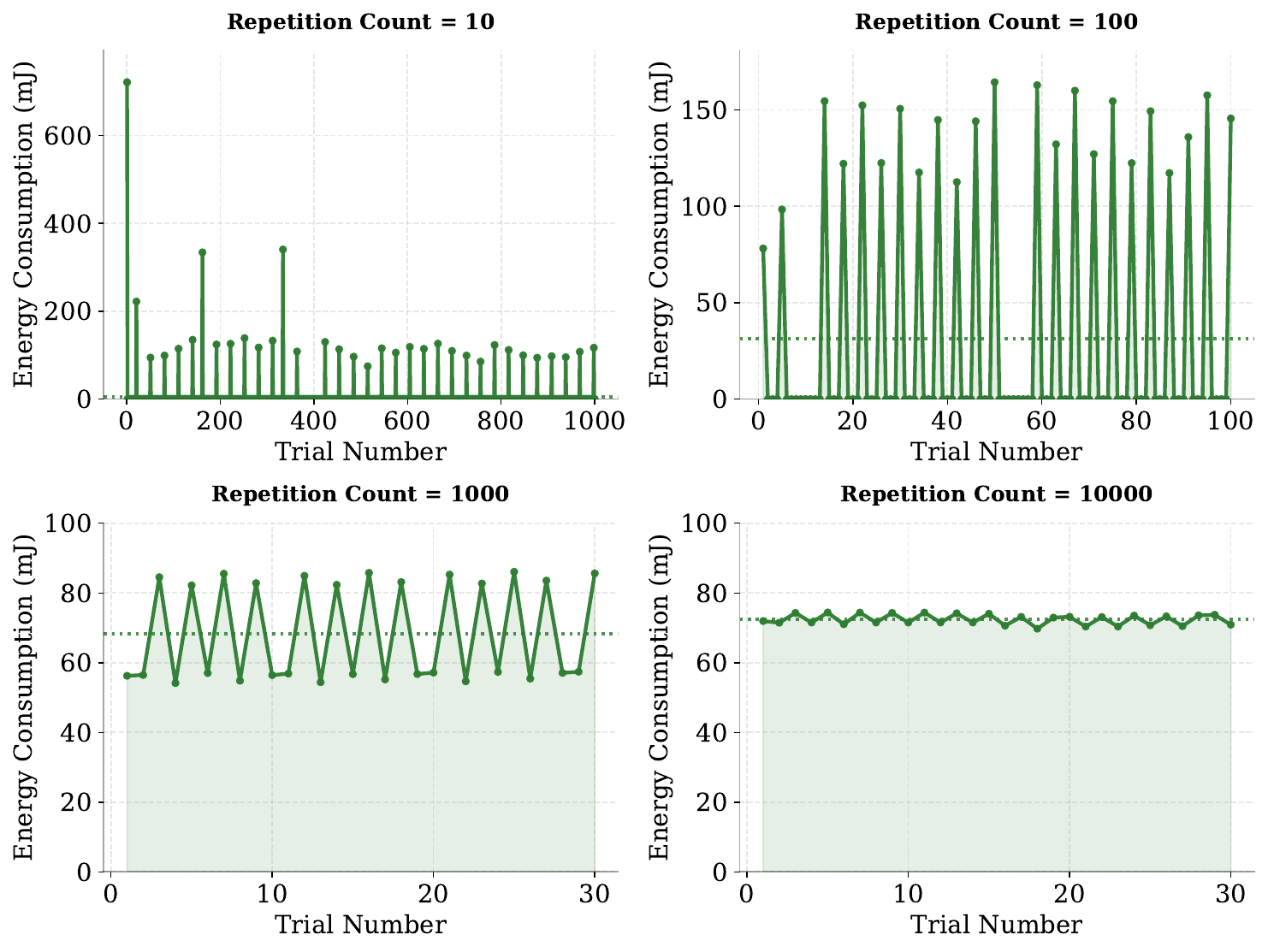}
    \caption{Effect of repeated sampling on energy measurements of the MLP block in Llama3.2-3B. As the repetition count increases, the variance in measured energy across trials decreases, indicating improved  reliability of energy measurements.}
    \label{fig:sampling}
    \vspace{-13pt}
\end{figure}

%% file: Files/ExperimentalDetails.tex
\subsection{Hyperparameter \& Hardware}

As part of our experimental protocol, we evaluate two floating-point precisions, FP32 and FP16, while varying the input sequence length across 8, 32, 64, 128, 256 and 512 tokens to study the scaling effects. Each configuration is run for a fixed set of 20 trials ($T = 20$) to capture variability and validate precision. We assume FP16 precision unless explicitly stated. 

We also investigate the effect of the repetition count $N$ on reliability of the energy readings. As seen in Fig. \ref{fig:sampling}, increasing repetition count ($N$) yields more stable readings and reduces measurement failures i.e. cases where the recorded energy for a trial is spuriously 0 mJ due to sensor granularity.
Based on the analysis, we set the repetition count $N=10,000$ for measurements of small components with execution time of order 100 $\mu$ s  and $N=1,000$ for energy measurements of the full model, balancing measurement accuracy with computational cost.

Experiments were conducted on NVIDIA Ada-Lovelace GPUs (RTX 6000 Ada \cite{NVIDIA-RTX6000AdaDatasheet}). The NVML interface typically updates power readouts only every 20–50 ms 
~\citep{Yang_2024,nik2025energyconsciousllmdecodingimpact} and introduced limitations in resolving microsecond-scale component execution, thereby requiring \textbf{\textcolor{CLEAR}{CLEAR}} for fine-grained energy attribution. Refer Appendix \ref{app:hardware} for details.

\input{Tables/GPT-OSS}

\subsection{Metrics}

% \subsubsection{Energy Consumption}
The energy consumed by each model component is measured in milliJoules(mJ), matching the $\approx$0.8mJ precision of the NVML sensor used. For validating our methodology (\circLabel{ii}), we define two complementary metrics, \textbf{Energy Captured} (Capture) and \textbf{Percentage Capture} (\%Capture). 
Energy Captured (in mJ) represents the total energy measured across all the major components of a given layer block or the entire model. 
Due to the limited precision of the instrumentation, we neglect negligible contributors (e.g., residual connections) and introduce \%Capture to indicate how well \textbf{\textcolor{CLEAR}{CLEAR}} accounts for the model's overall energy usage. 
Specifically, \%Capture is the ratio of the measured Energy Captured to the model's measured energy consumption, expressed as a percentage:
\[
\begin{aligned}
\text{Capture} &= \sum_{i=1}^{N} \bar{E}_i , \quad
\% \text{Capture} = \frac{\sum_{i=1}^{N} \bar{E}_i}{\bar{E}_{\text{model}}} \times 100
\end{aligned}
\]
% where \(\bar{E}_i\) denotes the energy (in mJ) consumed by the \(i^{\text{th}}\) major component and \(E_{\text{total}}\) is the total energy consumption of the block or entire model.

% \subsubsection{Computational Efficiency}
 \textbf{\textcolor{CLEAR}{CLEAR}} quantifies the computation executed by each component using FLOPs and measures GPU execution time ($\mu$s) using the PyTorch Profiler. To evaluate the energy cost per unit of computation, we define two metrics:

\[
\text{E/FLOP} =
\frac{\text{Energy}\;[\mathrm{mJ}]}{\mathrm{FLOPs}\times 10^{-9}} 
\]
\[
{\Delta E}/{\Delta \text{FLOP}} =
\frac{\Delta Energy\;[\mathrm{mJ}]}{\Delta \mathrm{FLOPs}\times 10^{-9}}.
\]

The metric E/FLOP (expressed in mJ/GFLOP), represents the average energy cost per unit computation, where lower values indicate higher energy efficiency. The marginal metric $\Delta E/\Delta\text{FLOP}$ measures the additional energy consumed per extra GFLOP, capturing the sensitivity of energy consumption to increased computational demand.

For details about list of models and transformer architectures, Refer Appendices \ref{app:Models} and \ref{app:Transformers}

% \subsection{Models}
% To conduct a systematic study of energy consumption across architectural paradigms,
% we consider four representative classes of Transformer-based models namely \textbf{Encoder-
% only} models, \textbf{Decoder-only} models, \textbf{Encoder-Decoder} models and sparse-activated \textbf{Mixture of Experts} (MOE) variants. Our model selection aims to balance breadth of architectural diversity with controlled comparisons of scale and design choices. 

% We evaluate eight widely-used Encoder only models namely BERT-base, BERT-large ~\citep{bert}, ALBERT-base, ALBERT-large~\citep{albert}, RoBERTa-base, RoBERTa-large~\citep{roberta} and distilled variants DistilBERT and DistilBERT~\citep{distilbert}. Base and large versions allow us to isolate the effect of model size on energy usage where distilled counterparts enable comparison with lightweight compression techniques. To represent contemporary LLMs i.e. Decoder-only models, we experiment with instruction-tuned variants of four key open-source families namely LLaMA 3.2-3B~\citep{llama3modelcard}, Gemma 3-4B~\citep{gemma}, Qwen 2.5-3B~\citep{qwen2}, and Phi-4-4B~\citep{abdin2024phi4technicalreport}. We focus specifically on single-token generation to control for variability in output sequence length and to minimize cache based auto-regressive generation. We also evaluate \textbf{\textcolor{CLEAR}{CLEAR}} on two well-established sequence-to-sequence models, namely BART~\citep{bart} and FLAN-T5~\citep{flant5} and a sparse-activated MoE, GPT-OSS-20B~\citep{gptoss} 

%% file: Tables/GPT-OSS.tex
% \usepackage{multirow}
\begin{table*}[]
\resizebox{\textwidth}{!}{%
\renewcommand{\arraystretch}{1.3}
\begin{tabular}{|ccccccccccc|}
\hline
\multicolumn{1}{|c|}{Component} & \multicolumn{2}{c|}{8 Tokens}                              & \multicolumn{2}{c|}{32 Tokens}                             & \multicolumn{2}{c|}{64 Tokens}                             & \multicolumn{2}{c|}{96 Tokens}                             & \multicolumn{2}{c|}{128 Tokens}       \\ \cline{2-11} 
\multicolumn{1}{|c|}{}                           & \multicolumn{1}{c|}{Avg.} & \multicolumn{1}{c|}{Std. Dev.} & \multicolumn{1}{c|}{Avg.} & \multicolumn{1}{c|}{Std. Dev.} & \multicolumn{1}{c|}{Avg.} & \multicolumn{1}{c|}{Std. Dev.} & \multicolumn{1}{c|}{Avg.} & \multicolumn{1}{c|}{Std. Dev.} & \multicolumn{1}{c|}{Avg.} & Std. Dev. \\ \hline                        
\multicolumn{1}{|c|}{Attention Block}            & 53.261                    & \multicolumn{1}{c|}{1.677}     & 64.147                    & \multicolumn{1}{c|}{0.686}     & 75.161                    & \multicolumn{1}{c|}{0.76}      & 93.91                     & \multicolumn{1}{c|}{0.779}     & 100.701                   & 1.045     \\
\multicolumn{1}{|c|}{MLP}                        & 685.408                   & \multicolumn{1}{c|}{12.61}     & 776.905                   & \multicolumn{1}{c|}{3.166}     & 867.687                   & \multicolumn{1}{c|}{0.867}     & 958.134                   & \multicolumn{1}{c|}{1.406}     & 1046.2                    & 1.187     \\
\multicolumn{1}{|c|}{Norm. (All)}                & 9.324                     & \multicolumn{1}{c|}{0.729}     & 10.787                    & \multicolumn{1}{c|}{0.825}     & 12.702                    & \multicolumn{1}{c|}{1.056}     & 13.443                    & \multicolumn{1}{c|}{1.422}     & 14.639                    & 1.108     \\ \hline
\multicolumn{1}{|c|}{Captured (Block)}           & 747.993                   & \multicolumn{1}{c|}{-}         & 851.839                   & \multicolumn{1}{c|}{-}         & 955.55                    & \multicolumn{1}{c|}{-}         & 1065.487                  & \multicolumn{1}{c|}{-}         & 1161.541                  & -         \\
\multicolumn{1}{|c|}{Block}                      & 731.905                   & \multicolumn{1}{c|}{12.456}    & 856.309                   & \multicolumn{1}{c|}{1.428}     & 951.869                   & \multicolumn{1}{c|}{0.805}     & 1057.01                   & \multicolumn{1}{c|}{0.881}     & 1157.197                  & 1.181     \\
\multicolumn{1}{|c|}{\% Capture (Block)}         & 102.198                   & \multicolumn{1}{c|}{-}         & 99.478                    & \multicolumn{1}{c|}{-}         & 100.387                   & \multicolumn{1}{c|}{-}         & 100.802                   & \multicolumn{1}{c|}{-}         & 100.375                   & -         \\ \hline
\multicolumn{1}{|c|}{Embedding Layer}            & 0.568                     & \multicolumn{1}{c|}{0.215}     & 0.627                     & \multicolumn{1}{c|}{0.282}     & 1.061                     & \multicolumn{1}{c|}{0.41}      & 1.077                     & \multicolumn{1}{c|}{0.434}     & 0.766                     & 0.357     \\
\multicolumn{1}{|c|}{LM Head}                    & 443.391                   & \multicolumn{1}{c|}{1.108}     & 452.139                   & \multicolumn{1}{c|}{0.988}     & 460.383                   & \multicolumn{1}{c|}{0.988}     & 475.22                    & \multicolumn{1}{c|}{1.265}     & 483.515                   & 1.242     \\
\multicolumn{1}{|c|}{Final Layer Norm.}          & 4.695                     & \multicolumn{1}{c|}{0.368}     & 5.14                      & \multicolumn{1}{c|}{0.361}     & 6.071                     & \multicolumn{1}{c|}{0.496}     & 6.625                     & \multicolumn{1}{c|}{0.525}     & 7.221                     & 0.466     \\ \hline
\multicolumn{1}{|c|}{Captured (Model)}           & 18014.38                  & \multicolumn{1}{c|}{-}         & 21009.32                  & \multicolumn{1}{c|}{-}         & 23312.36                  & \multicolumn{1}{c|}{-}         & 25851.15                  & \multicolumn{1}{c|}{-}         & 28264.24                  & -         \\
\multicolumn{1}{|c|}{Model}                      & 18447.5                   & \multicolumn{1}{c|}{63.784}    & 21366.69                  & \multicolumn{1}{c|}{103.479}   & 24126.47                  & \multicolumn{1}{c|}{12.67}     & 26634.05                  & \multicolumn{1}{c|}{15.33}     & 28801.98                  & 2.867     \\
\multicolumn{1}{|c|}{\% Capture (Model)}         & 97.652                    & \multicolumn{1}{c|}{-}         & 98.327                    & \multicolumn{1}{c|}{-}         & 96.626                    & \multicolumn{1}{c|}{-}         & 97.061                    & \multicolumn{1}{c|}{-}         & 98.133                    & -         \\ \hline
\end{tabular}
}
\caption{Energy Consumption for GPT-OSS-20B model across different input token length with \%Capture (96+\%) for Block and Full Model and Std. Deviation of Energy Consumed across 20 trials. All energy units in mJ.}
\vspace{-12pt}
\label{tab:gpt-oss}
\end{table*}

%% file: Files/Results.tex
\input{Figures/batch}
\input{Figures/multi-token}
\label{Results}
\textit{Completeness of Energy Captured}: Despite the omission of very small and negligible components, the overall \%Capture at both the block and model level consistently remains above 90\% across different models (Refer Table \ref{tab:gpt-oss} and Appendix \ref{app:Results}). As per \circLabel{ii}, we observe that energy of individual components provide a reliable and near-complete estimate of the total energy consumption dictated by the model’s architecture.
However we consistently observe low \%Capture(Block) of ALBERT variants possible due to factorized embeddings causing higher idle energy consumption.\\
\textit{Consistency Across Trials} : Using \textbf{\textcolor{CLEAR}{CLEAR}}, we observe that the average standard deviation of the measured component energies consistently remains below 9.5\% of the respective mean for components consuming $>5$mJ of energy. As component size and execution time increase, the relative standard deviation decreases(eg: components consuming approximately 1J show deviations as low as 1\% \circLabel{i}). Such behavior arises because shorter executions yield fewer sensor samples, making measurements more sensitive to idle-energy noise and the sensor’s precision limits (Refer Appendix \ref{app:StdDev}.)

\subsection{Batch Size}
Empirically, we observe that increasing the batch size leads to a substantial reduction in per-sample energy consumption. \textbf{\textcolor{CLEAR}{CLEAR}} investigates the effect of individual model components in overall energy savings. Even a modest increase to a batch size of 2 results in a significant reduction of approximately 40–45\% (See Figure \ref{fig:batch}) in per-sample energy across major components. The reduction is slightly more pronounced in the MLP blocks compared to the Attention blocks due to higher arithmetic intensity and more efficient utilization of GPU compute units. At larger batch sizes, the per-sample energy consumption decreases by up to 80\%. These results indicate that batching is a critical optimization not only at the system level but also within the internal components of the model.

\subsection{Impact of KV Cache}
Using \textbf{\textcolor{CLEAR}{CLEAR}}, we extend the analysis beyond single-token \textit{Prefill} stage to study realistic multi-token generation in Decoder-only Transformer models. In autoregressive \textit{Decode} stage, each generated token attends to all previously processed tokens. Without optimization, decoding requires recomputing the Key ($K$) and Value ($V$) projections for the entire sequence at every generation step, leading to substantial increase in computational cost and energy consumption as the sequence length grows. Key–Value (KV) cache eliminates the computational redundancy by storing previously computed $K$ and $V$ tensors during the \textit{Prefill} stage and reusing them during subsequent \textit{Decode} steps. During decoding, only the Query ($Q$) for the newly generated token must be computed, while all previous keys and values are retrieved directly from the cache, reducing both computation and memory movement. (Refer Appendix \ref{app:MultiToken} for more details)

Empirically, we observe that KV Cache has a substantial impact on energy efficiency. When KV cache is enabled, both the FLOPs ($\sim$0.019 GFLOPs) and energy consumption of the attention mechanism remain nearly constant across increasing sequence lengths. In contrast, disabling KV cache results in rapid increase of computational cost, with FLOPs rising to 9.667 GFLOPs and energy increasing from 28.05 mJ to 87.36 mJ as the input sequence grows from 1 to 512 tokens (See Figure \ref{fig:multitoken} \circLabelMULT{b}). As seen in \circLabelMULT{a}, KV cache reduces Energy consumption by more than 60\% at longer contexts whereas the energy gains are negligible for shorter sequences of about 50 input tokens.

\subsection{Attention Variants \& Optimizations}
\input{Figures/variant}

\input{Figures/gemma}
\input{Figures/dimension}
\textbf{\textcolor{CLEAR}{CLEAR}}, with its Amplification Strategy, enables fine-grained energy analysis of different component variants and optimizations. We compare three Attention implementations: \textit{Eager Attention}, \textit{Scaled Dot-Product Attention (SDPA)}, and \textit{Flash Attention}. Across both models, Flash Attention consistently consumes less energy than Eager Attention, while SDPA falls in between. This occurs because Eager Attention materializes the full attention matrix, increasing memory traffic and kernel launches whereas Flash Attention computes attention in tiled blocks without storing the complete matrix, reducing memory movement, and improving energy efficiency (Refer Figure \ref{fig:variant})

% Table~\ref{Attention-Optimization} reports the energy consumption for Qwen2.5-3B and Gemma3-4B across input lengths of 64, 128, and 256 tokens.
Applying \textbf{torch.compile()} further reduces energy consumption for both Flash and Eager implementations. The improvement comes from graph-level optimizations that fuse multiple small kernels into larger kernels, reducing kernel launch overhead and improving GPU utilization. Refer Appendix \ref{app:Variants} for more details about the optimizations.
The largest reduction in energy is observed with the \textbf{Max Autotune} and \textbf{Reduced Overhead} optimizations, as they aggressively fuse operations and remove runtime overheads such as profiling and synchronization, resulting in more efficient execution paths with fewer intermediate operations.

\subsection{Hidden Dimension \& Attention Heads}
% As shown in Figure \ref{fig:dimension}, we observe that MLP blocks scale perfectly linearly with hidden dimension size with a very small constant overhead (about 0.7mJ per 100 dimension) whereas for a 256 hidden dimension Attention block, the constant is about 31.34 mJ and it scales with 3.27 mJ per attention head. The relatively higher contant for Attention can be explained by ...
As shown in Figure~\ref{fig:dimension}, we observe that Energy Consumed by the MLP block scales almost perfectly linearly with the hidden dimension size, with only a very small constant overhead (approximately $0.7\,\mathrm{mJ}$ per 100 hidden dimensions). The behavior is expected because the MLP is primarily composed of dense matrix multiplications whose computational cost increases proportionally with the hidden dimension. In contrast, the Attention block exhibits a substantially larger constant overhead. For a configuration with 256 hidden dimensions, the constant energy cost is approximately $31.34\,\mathrm{mJ}$, and the energy consumption increases by about $3.27\,\mathrm{mJ}$ per additional Attention Head.

Above observations have important architectural implications. Increasing the hidden dimension mainly affects the MLP energy cost and scales in a predictable linear manner. In contrast, increasing the number of attention heads introduces both a significant baseline cost and an additional per-head overhead, making attention scaling relatively more energy expensive.

% \vspace{-5pt}
\subsection{FLOPs ~ \& ~ Energy}
\label{Flops}
As shown in Figure \ref{fig:gemma} \circLabelMULT{b}, across all input token lengths, the Attention mechanism consistently exhibits a higher Energy/FLOP ratio than MLP layers, LM-Head and overall model. MLP and LM-Head layers consist dense matrix multiplications and are efficiently accelerated by GPUs, executing more computations per unit of energy. Attention, however, involves multiple stages like query–key dot product, softmax operations, and irregular memory accesses that can introduce additional memory and synchronization overheads. 

As shown in figure \ref{fig:gemma} \circLabelMULT{a}, we consistently observe that E/FLOP ratio decreases as the input sequence length increases for all components. When longer sequences are processed, the fixed costs associated with computation are amortized over a larger number of tokens. To better isolate the energy associated with additional computation, we analyze the marginal energy cost per FLOP, defined as $\Delta E/\Delta \text{FLOP}$. The marginal energy remains approximately constant as input length increases across model components indicating a strong empirical evidence that FLOPs are the primary driver of the \textbf{variable} portion of energy consumption and can be decomposed as 
\begin{equation}
  E(L) \approx E_{0} + k \cdot \mathrm{FLOPs}(L)
  \label{eq:model}
\end{equation}
where $E_0$ denotes a fixed energy overhead and $k$ represents the marginal energy cost per FLOP.  \(L\) denotes input length and only the second term grows nearly linearly with the
computational workload (FLOPs). $k$ is component-dependent constant of proportionality and is noticeably higher for the Attention mechanism. Overall results indicate that while FLOPs explain the variable component of energy consumption, accurate component-level energy estimation must account for fixed overheads ($E_0$) and component-specific marginal costs ($k$).

%% file: Figures/batch.tex
\begin{figure}[t]
    \centering
    \includegraphics[width=\columnwidth]{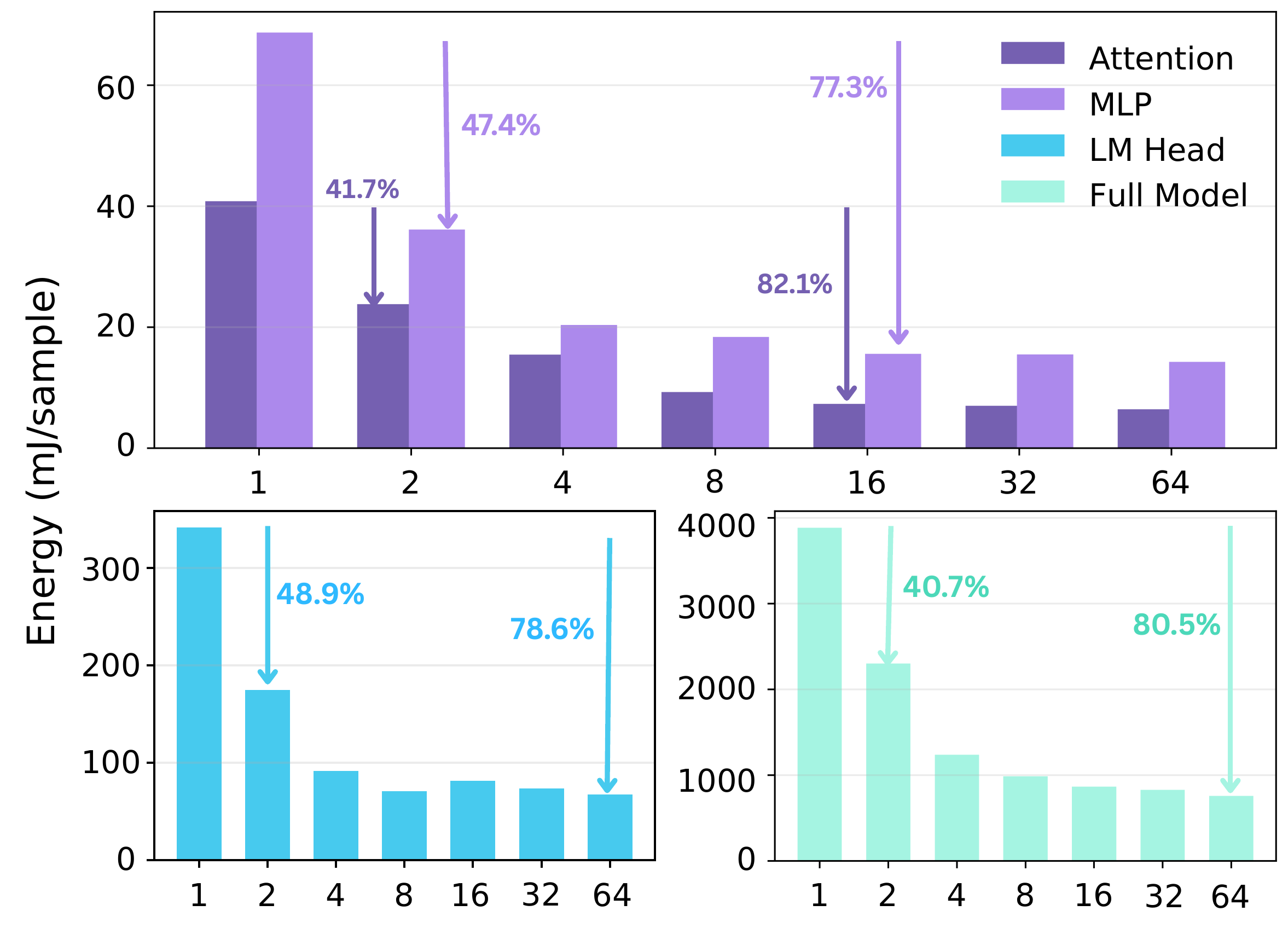}
    \caption{Energy consumed per sample by different components of Llama-3.2-3B as a function of batch size. Increasing the batch size significantly improves energy efficiency. X-axis represents the Batch size. }
    \label{fig:batch}
    \vspace{-12pt}
\end{figure}

%% file: Figures/multi-token.tex
\begin{figure*}[t]
    \centering
    \includegraphics[width=0.98\textwidth]{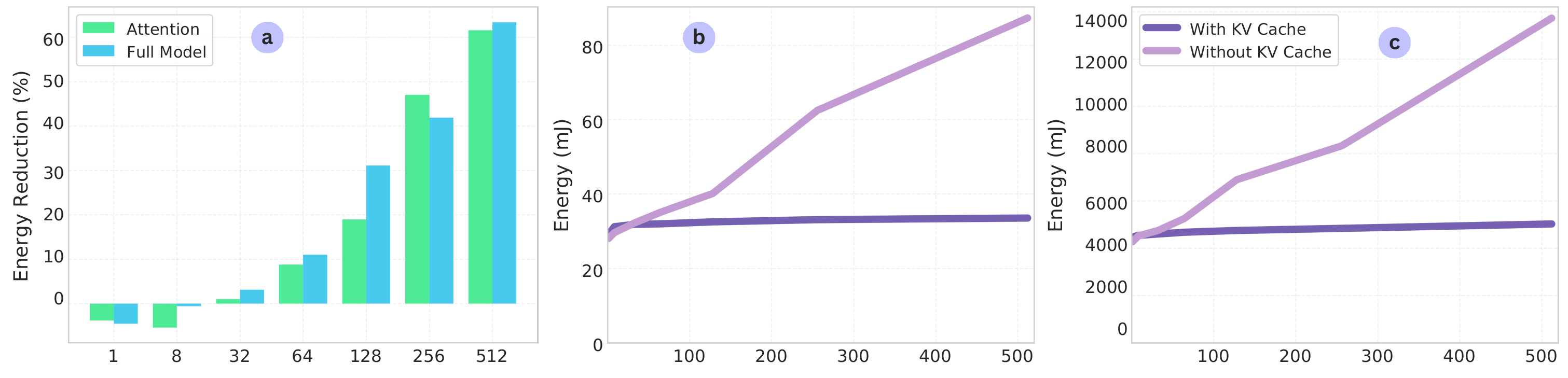}
    \vspace{-8pt}
    \caption{(a) Percentage reduction in energy due to KV cache for Attention and full model across different input sequence lengths. (b) and (c) show the energy consumption of the Attention block and full model, respectively, as a function of input token length during single-token generation. Experiments are performed on Qwen2.5-3B.}
    \label{fig:multitoken}
    \vspace{-13pt}
\end{figure*}

%% file: Figures/variant.tex
\begin{figure}[t]
    \centering
    \includegraphics[width=\columnwidth]{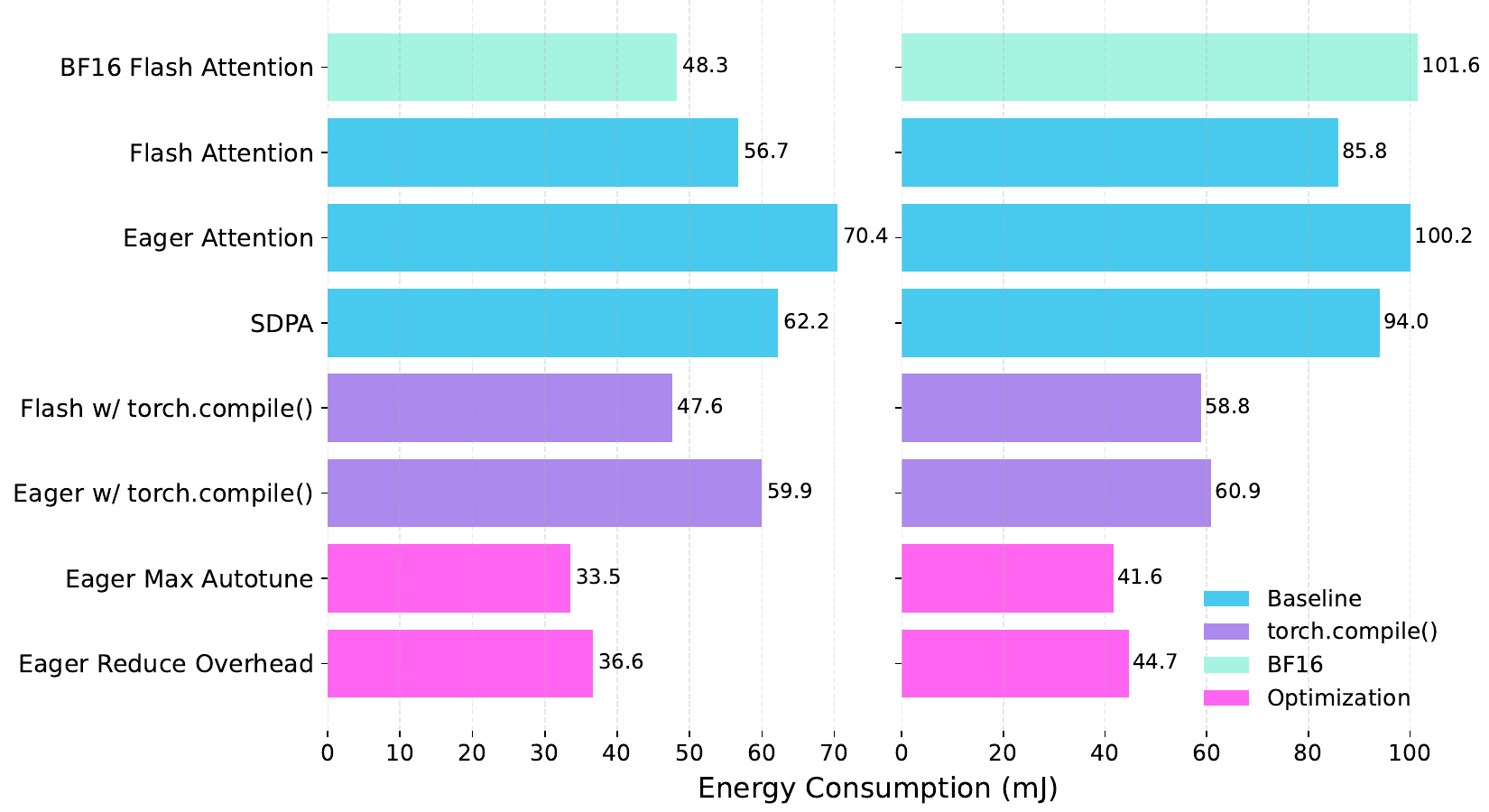}
    \caption{Energy Consumption for different variants and optimizations of Attention Mechanism on Qwen2.5-3B (left) and Gemma3-4B (right) on 256 input tokens}
    \label{fig:variant}
    \vspace{-13pt}
\end{figure}

%% file: Figures/gemma.tex
\begin{figure*}[t]
    \centering
    \includegraphics[width=0.98\textwidth]{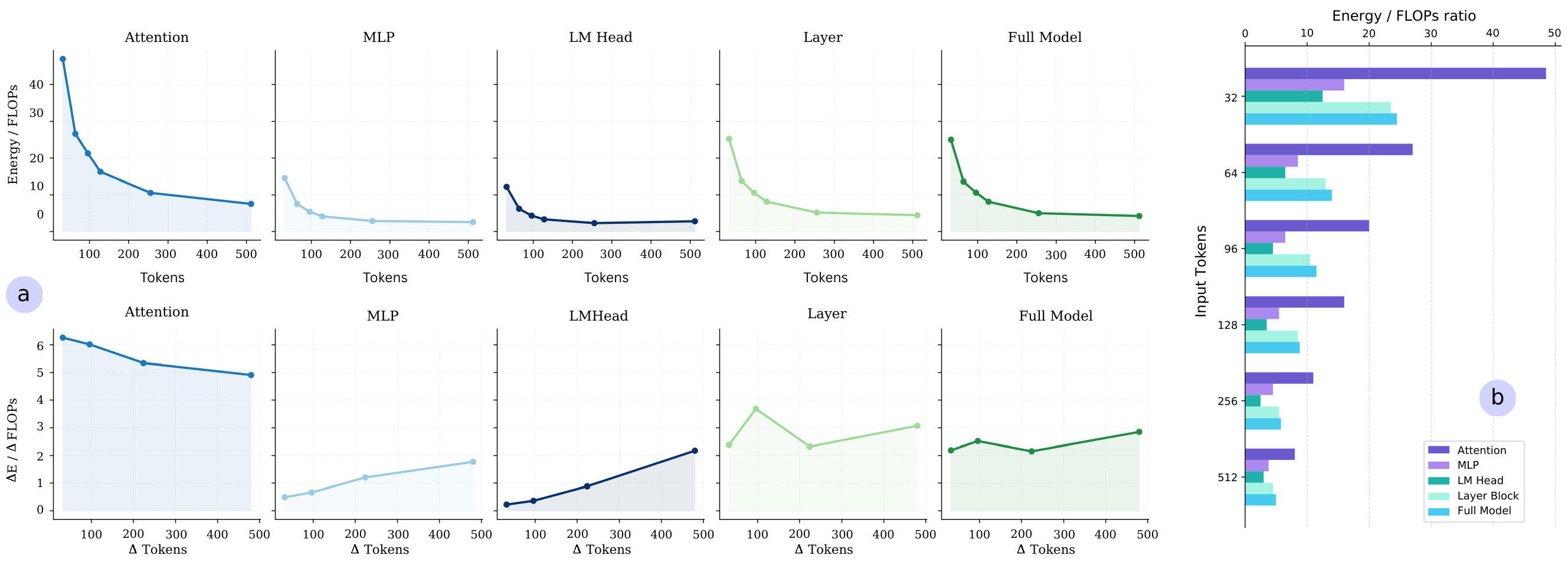}
    \vspace{-9pt}
    \caption{ (a) Variation of $E/\text{FLOP}$ and $\Delta E/\Delta \text{FLOP}$ for Gemma3-4B model across major components. E/FLOP ratio decreases with input sequence length, while marginal energy consumed per FLOP remains nearly constant. (b) Attention consistently shows the highest E/FLOP ratio across all input lengths.}
    \label{fig:gemma}
    \vspace{-10pt}
\end{figure*}

%% file: Figures/dimension.tex
\begin{figure}[t]
    \centering
    \includegraphics[width=\columnwidth]{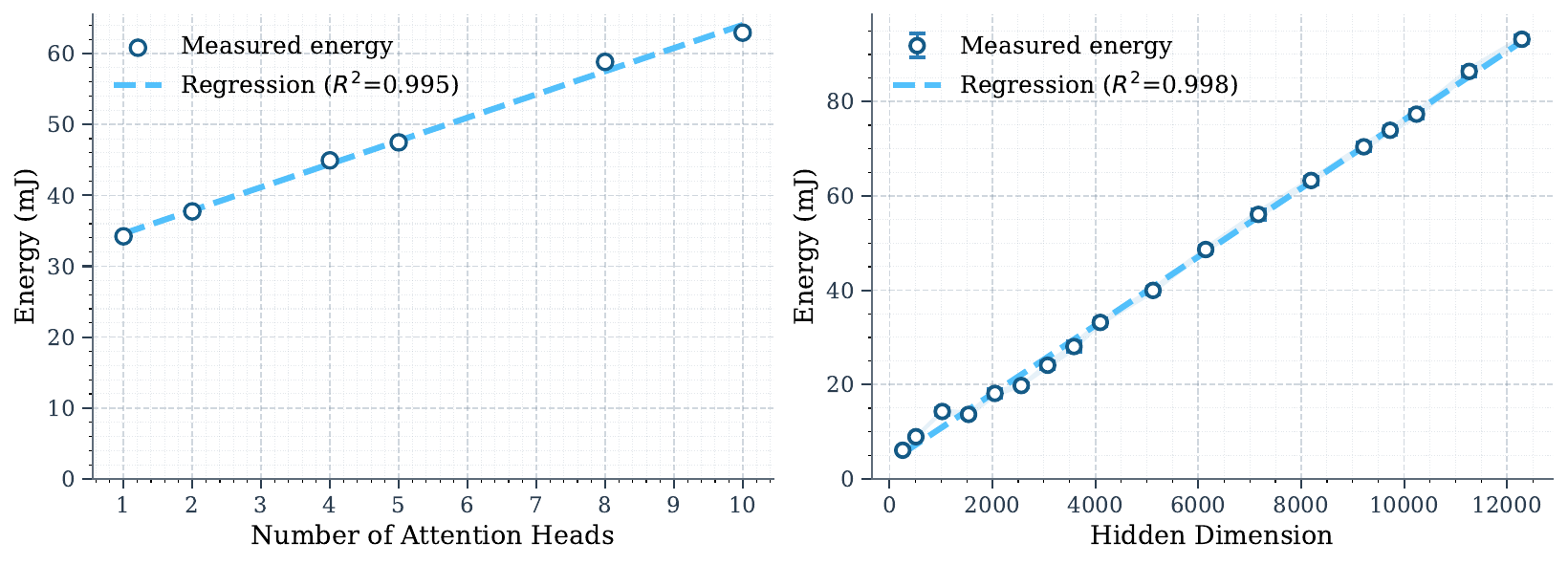}
    \caption{Energy Scaling for Attention and MLP block as a function of Number of Attention Heads (left) and Hidden Dimension (right) respectively.}
    \label{fig:dimension}
    \vspace{-18pt}
\end{figure}

%% file: Files/Discussion.tex
Most sustainability studies have primarily focused on model-level energy consumption, treating LLMs as \emph{monoliths} and paying little attention to the heterogeneity of their internal components from a sustainability perspective. \textbf{\textcolor{CLEAR}{CLEAR}}'s contribution to measure energy consumption at component-level granulity has direct implications for the research community as it provides a systematic methodology to reliably measure internal energy dynamics and enable targeted energy optimizations at the design level of model architecture. 

Previous works \cite{cao2021ireneinterpretableenergyprediction,10793163,vandervlugt2025powersensor3fastaccurateopen, s2024cppjoulesenergymeasurementtool} do measure energy consumption using the two paradigms: \circLabelPink{a} Using large datasets and model-level granularity \circLabelPink{b} Hardware-based approach. However, the above works explicitly acknowledge a key limitation: the coarse sampling frequency of available GPU/CPU energy sensors fundamentally restricts the granularity at which Energy consumption can be attributed at the software scale. As a result, they are unable to isolate the actual energy usage of microsecond-scale sub-operations or validate whether the reported estimates precisely reflect true energy consumption. \textbf{\textcolor{CLEAR}{CLEAR}}’s three-stage pipeline is hardware agnostic as it runs at an application/software level which allows easy adaptability for fine-grained energy analysis in Transformer architectures.

As discussed in Section \ref{Flops}, we observe that each component consumes energy disproportionately, posing a threat to use FLOPs and related metrics as convenient proxies for energy consumption \citep{getzner2023accuracymetricmattersestimating, ozcan2025quantifyingenergyconsumptioncarbon} as component-level disparities are systematically obscured by model-level aggregate measurements.

As seen in Section \ref{Results}, \textbf{\textcolor{CLEAR}{CLEAR}} demonstrates statistically reliable and sufficiently complete component-wise energy profiling that can be employed to support comparative energy analysis and draw robust conclusions about energy implications of specific design choices. \textbf{\textcolor{CLEAR}{CLEAR}} establishes a foundation for future work on predictive modeling so that the energy costs can be computed based on architectural design choices like hidden dimensions, number of layers, etc., allowing accurate, generalizable prediction of component-wise energy in the early design stages. This aligns with the growing emphasis on Green AI and the need for energy-aware, sustainable AI system design. \citep{greenAIstudy,greenAI2,greenAI3}

\textbf{\textcolor{CLEAR}{CLEAR}} provides a strong formal foundation for predictive modeling of energy consumption and enables the recording of reliable and accurate ground-truth measurements. Previous works like IrEne \citep{cao2021ireneinterpretableenergyprediction} use a regression-based prediction model on feature dimensions to predict the energy consumption of the model. \textbf{\textcolor{CLEAR}{CLEAR}} can enhance IrEne's energy prediction approach by providing large-scale, reliable ground truths and insights on the impact of different parameters on energy for better predictive modeling.

\section{Conclusion}
Taken together, \textbf{\textcolor{CLEAR}{CLEAR}}'s findings underscore that sustainability in AI must be treated as  \textbf{a first-class research objective rather than an afterthought}. By moving beyond aggregate model-level reporting to examine component-level dynamics, we aim to motivate the software and AI research communities to pursue progress that is both holistic and environmentally responsible, driven by a proactive rather than reactive mindset. Looking forward, we hope this work inspires future research to integrate energy considerations into every stage of model development, fostering AI systems that are not only performant but also sustainable.

%% file: Files/Ending.tex
\section{Limitations}
\label{limitations}

While our study provides the first component-level view of energy consumption in Transformer architectures, a few aspects merit further exploration. 

First, our energy estimates rely on NVIDIA’s NVML interface and FLOP counts obtained via the PyTorch Profiler. Though ther are well-established tools, but like all measurement frameworks they may carry some inherent uncertainties that may slightly affect the results by negligible margin.

Second, GPU hardware introduces additional variability. Different GPU families and generations apply their own low-level optimization, potentially affecting the energy profiles of specific computational components. Extending this analysis across a wider range of hardware would sharpen our understanding of how these optimizations influence component-wise energy usage. While it is left for future work, to the best of our knowledge, our study remains the first systematic investigation of component-level energy dynamics in Transformer models, providing a foundational understanding of the challenges and opportunities present on current hardware.

Finally, because prior literature offers little precedent for fine-grained energy measurement of individual Transformer components, our validation protocol represents an initial step. Future work can strengthen and expand these validation methods as the research community develops more sophisticated benchmarks and measurement standards.

\section*{Ethical Considerations}
This work studies the energy consumption of Transformer-based models at a fine-grained component level to better understand and improve the environmental efficiency of large language model inference. The proposed \textbf{\textcolor{CLEAR}{CLEAR}} framework measures energy consumption of internal model components under controlled experimental conditions using publicly available models and standard hardware instrumentation.

Our work does not involve human subjects, personal data, or sensitive information. While the methodology could be used to optimize systems for energy efficiency, we emphasize that such optimizations should be considered alongside other AI considerations such as safety, fairness, and reliability. Overall, our work aims to support the development of more sustainable and environmentally responsible AI systems.

% \textbf{\textcolor{CLEAR}{CLEAR}} allows

%% file: Appendix/Models.tex
\section{Models}
\label{app:Models}
To conduct a systematic study of energy consumption across architectural paradigms,
we consider four representative classes of Transformer-based models namely \textbf{Encoder-
only} models, \textbf{Decoder-only} models, \textbf{Encoder-Decoder} models and sparse-activated \textbf{Mixture of Experts} (MOE) variants. Our model selection aims to balance breadth of architectural diversity with controlled comparisons of scale and design choices. 

We evaluate eight widely-used Encoder only models namely BERT-base, BERT-large ~\citep{bert}, ALBERT-base, ALBERT-large~\citep{albert}, RoBERTa-base, RoBERTa-large~\citep{roberta} and distilled variants DistilBERT and DistilBERT~\citep{distilbert}. Base and large versions allow us to isolate the effect of model size on energy usage where distilled counterparts enable comparison with lightweight compression techniques. To represent contemporary LLMs i.e. Decoder-only models, we experiment with instruction-tuned variants of four key open-source families namely LLaMA 3.2-3B~\citep{llama3modelcard}, Gemma 3-4B~\citep{gemma}, Qwen 2.5-3B~\citep{qwen2}, and Phi-4-4B~\citep{abdin2024phi4technicalreport}. We focus specifically on single-token generation to control for variability in output sequence length and to minimize cache based auto-regressive generation. We also evaluate \textbf{\textcolor{CLEAR}{CLEAR}} on two well-established sequence-to-sequence models, namely BART~\citep{bart} and FLAN-T5~\citep{flant5} and a sparse-activated MoE, GPT-OSS-20B~\citep{gptoss} 

%% file: Appendix/Transformers.tex
\section{Transformer Architectures}
\label{app:Transformers}

Despite architectural differences, the transformer based models share a set of common computational primitives. The detailed flow of how the model produces its output is illustrated in the following.

Let $\mathcal{V}$ denote the vocabulary space, and $x = (x_1, \ldots, x_T)$, $x_t \in \mathcal{V}$ be an input token sequence of length $T$. A tokenizer $\mathcal{T} : \mathcal{V} \rightarrow \{1,\ldots,|\mathcal{V}|\}$ maps tokens to discrete indices. The indices are embedded into a continuous space by the \textbf{Embedding Layer} $E \in \mathbb{R}^{|\mathcal{V}| \times d}$:
\[
h^0_t = E[x_t] + P_t, \quad t=1,\ldots,T,
\]

where $P_t \in \mathbb{R}^d$ denotes the positional embedding and $d$ is the hidden dimensionality. Embeddings are then passed to a set of Transformer layers each consisting of an \textbf{Attention} (Attn) block followed by a position-wise Feed-Forward \textbf{MLP} block, interleaved with residual connections and \textbf{Normalization} blocks for numerical stability, as depicted in the Transformer block of Figure \ref{fig:methodology}. For layer $\ell \in \{1,\ldots,L\}$, the computations are:

\[
\tilde{h}^\ell = \mathrm{LN}(h^{\ell-1}), \quad z^\ell = h^{\ell-1} + \mathrm{Attn}(\tilde{h}^\ell)
\]
\[ 
h^\ell = z^\ell + \mathrm{FFN}(\mathrm{LN}(z^\ell)).
\]

However, the placement of Normalization blocks vary across different models and can be applied at different stages within a layer block. Post all layers, the final hidden state $h^L \in \mathbb{R}^{T \times d}$ is projected onto the vocabulary using the language modeling head:
\[
\hat{y}_t = \mathrm{softmax}(h^L_t W_{\text{LM}}^\top), \quad W_{\text{LM}} \in \mathbb{R}^{|\mathcal{V}| \times d}.
\]

\subsection{Encoder-Only \label{subsec:encoder-only}}

Encoder-only architectures (e.g., BERT, RoBERTa, AlBERT) compute contextualized token representations using bidirectional self-attention. They are commonly used for classification, token-level prediction and masked-language-modeling objectives. The encoder processes the full input \(x\) in parallel, producing \(H^{L} = (h^L_1,\dots,h^L_T)\in\mathbb{R}^{T\times d}\) which can be pooled or projected to task heads.

Encoder-only architectures typically use a \emph{dense attention pattern}, meaning that no causal mask is applied to restrict attention.  
Formally, the attention operation is defined as
\begin{equation}
  \mathrm{Attn}(Q,K,V) = \mathrm{softmax}\!\left(\frac{QK^{\top}}{\sqrt{d_k}}\right)V,
  \label{eq:attention}
\end{equation}

where every token can attend to every other token in the sequence.  
This design produces \emph{symmetric contextualization}, since information can flow bidirectionally across tokens.  
However, it comes with a computational cost of \(\mathcal{O}(T^2 d)\) per layer (due to the pairwise interactions between all tokens) and a memory cost of \(\mathcal{O}(T^2)\) for storing the attention weights.

Different tasks attach specialized output heads on top of the final hidden states \(h^L_t \in \mathbb{R}^d\). For \textbf{token-level classification}, each token representation is projected into the label space using a weight matrix \(W_{\mathrm{tok}}\).  
The predicted class distribution for token \(t\) is given by
\[
  \hat{y}_t = \mathrm{softmax}\!\left(h^L_t W_{\mathrm{tok}}^{\top}\right).
\]

For \textbf{sequence-level classification}, the hidden state of a special token such as \([{\rm CLS}]\) serves as a summary vector for the entire sequence.  
This representation \(h^L_{\mathrm{cls}}\) is then passed through a classifier, typically implemented as a multi-layer perceptron (MLP):
\[
  h^L_{\mathrm{cls}} \mapsto \mathrm{MLP}(h^L_{\mathrm{cls}}).
\]

For \textbf{masked language modeling (MLM)}, the prediction head reuses the input embedding matrix \(E\) to tie input and output representations.  
In this case, the output weight matrix is defined as
\[
  W_{\mathrm{LM}} = E.
\]
This weight sharing enforces consistency between how tokens are encoded and how they are predicted.

Encoder-only designs maximize parallelizability during training as the whole sequence is processed concurrently, but the \(T^2\) attention cost and the need to store full-layer activations drive both memory bandwidth and energy cost during training.

\input{Tables/Encoder-only-model-info}

\subsection{Decoder-Only (Autoregressive) \label{subsec:decoder-only}} 

Decoder-only (autoregressive) models (e.g. Llama3.1, GPT) perform next-token prediction and are optimized for generative tasks. The decoder processes tokens causally with a triangular mask in which each position can attend only to tokens at previous positions (and itself) to enforce autoregressive factorization. The causal mask \(M_{\mathrm{causal}}\) has entries \(0\) for allowed positions and \(-\infty\) for disallowed future positions, implementing the triangular attention.

\paragraph{Inference optimization: KV-caching.} During autoregressive generation, previously computed keys and values can be cached: for step \(t\) only the new query interacts with stored \(K_{1:t-1},V_{1:t-1}\). This reduces per-step attention cost from \(\mathcal{O}(t^2 d)\) to \(\mathcal{O}(t d)\) (amortized), and reduces the energy per generated token substantially at inference time. Training a decoder-only model still incurs full-sequence \(\mathcal{O}(T^2 d)\) attention cost, but inference benefits from KV-caching. Energy per generated token at inference depends on cache memory bandwidth and per-layer dot-product costs; thus memory movement for KV-cache and tiled attention matmuls can dominate measured energy.

 The set of activation hooks\(\mathcal{A}_{\mathrm{dec}}\) defined similarly to the encoder case, but is adapted to handle causal inputs and cached key-value states and it stores the intermediate activations \(\alpha^{\texttt{attn\_in}}_{\ell,t}\) and \(\alpha^{\texttt{kv}}_{\ell,t}\) for each layer \(\ell\) and time step \(t\). For profiling isolated attention at generation-time, replaying using cached KV tensors models the exact inference cost. For the purpose of our analysis, we primarily consider the energy and computation associated with the generation of \textbf{a single token}, where KV-caching is not utilized
 
Decoder-only models are majorly different from Encoder-only models as they prioritize autoregressive causality in which causal masking changes attention sparsity and reduces parallelism during sequence generation. KV-caching and rotary/relative positional encodings are often used to support long-context amortized inference and decoder-only models commonly use tied input/output embeddings to reduce parameter counts, and favor pre-norm residual stacks for stability in deep networks. Refer Tables \ref{tab:decoder_5000_GPU}, \ref{decoder-fp16-6000-tok}, \ref{decoder-fp16-6000-8tok}, \ref{decoder-fp32-6000-8tok}

\input{Tables/Decoder-only-model-info}

\subsection{Encoder-Decoder \label{subsec:encoder-decoder}}

Encoder-Decoder models, also known as sequence-to-sequence (Seq2Seq) architectures, are widely used for tasks requiring input-to-output transformations such as machine translation, summarization, and code generation. Formally, given an input sequence \(x = (x_1, \dots, x_{T_\text{in}})\), the encoder maps it to a sequence of hidden representations \({H} = (h_1, \dots, h_{T_\text{in}})\), and the decoder generates an output sequence \(y = (y_1, \dots, y_{T_\text{out}})\) autoregressively, conditioned on \({H}\).

The encoder is a stack of \(L_e\) Transformer layers that performs contextual embedding of the input tokens. Each layer typically consists of:
\begin{itemize}
    \item An \textbf{Attention} (Attn) mechanism that captures global dependencies within the input sequence i.e. for layer \(\ell\)
    \[
    \tilde{h}^{\ell-1} = \mathrm{LN}(h^{\ell-1}) \]
    \[
    h'^{\ell} = h^{\ell-1} + \mathrm{Attn}^{\ell}(\tilde{h}^{\ell-1}) \]
    \[
    h^{\ell} = h'^{\ell} + \mathrm{FFN}^{\ell}(\mathrm{LN}(h'^{\ell})).
    \]
    \item \textbf{Feedforward Network} that adds per-position nonlinear transformation to learn deeper features.
\end{itemize}
The encoder produces rich representations that capture semantic and syntactic relationships within the input sequence.

The decoder is also a stack of \(L_d\) Transformer layers, each consisting of:
\begin{itemize}
    \item \textbf{Masked self-attention} which ensures autoregressive generation by attending only to previous positions.
    \item \textbf{Encoder-decoder cross-attention} mechanism to attend to the encoder hidden states \({H}\), incorporating information from the entire input sequence into each decoding step.
    \item A \textbf{Feedforward network} similar to the encoder.
\end{itemize}
Mathematically, for decoder layer \(\ell\):
\[
\tilde{u}^{\ell-1} = \mathrm{LN}(u^{\ell-1}),\quad
s_\text{self}^{\ell} = \mathrm{Attn}_\text{causal}(\tilde{u}^{\ell-1}),
\]
\[
s_\text{cross}^{\ell} = \mathrm{Attn}_\text{cross}(\mathrm{LN}(u^{\ell-1}+s_\text{self}^{\ell}), {H}),
\]
\[
u^\ell = \mathrm{FFN}(\mathrm{LN}(u^{\ell-1}+s_\text{self}^{\ell}+s_\text{cross}^{\ell})).
\]

Compared to encoder-only models, the encoder-decoder architecture introduces a separate decoder stack with cross-attention, which enables output generation conditioned on the full input sequence. In contrast, encoder-only models produce fixed-length or token-level representations that are typically used for classification or embedding tasks, without any autoregressive generation.

\input{Tables/Decoder-5000}

When compared to decoder-only models, encoder-decoder architectures separate the input encoding from the output generation, whereas decoder-only models combine both within a single autoregressive stack. This separation allows the encoder to process the entire input sequence in parallel, improving training efficiency. Furthermore, in terms of residual and attention patterns, encoder-decoder models incorporate both self-attention in the decoder and cross-attention between the decoder and encoder outputs, whereas encoder-only and decoder-only architectures contain only a single attention mechanism.

\textbf{Energy Perspective: }The two-stack design of encoder-decoder models increases the total parameter count and memory footprint, resulting in higher energy consumption during training compared to encoder-only models for sequences of the same length. However, the input encoding phase can be fully parallelized across positions, and autoregressive decoder computation can benefit from caching mechanisms during inference, which partially reduces the per-token energy cost

\subsection{Mixture of Experts (MoE)}

Mixture-of-Experts (MoE) architectures extend standard Transformers by introducing conditional computation i.e. instead of activating all parameters for every input token, only a subset of "expert" networks is selected dynamically. This allows scaling model capacity substantially while keeping per-token computation and energy consumption manageable.

An MoE layer contains \(E\) independent feedforward networks, or Experts, each with parameters \(\theta_1,\dots,\theta_E\). For a given token representation \(h\in\mathbb{R}^d\), the computation is routed through a small subset of \(k < E\) experts, typically \(k=2\) or \(3\)

\[
m^\text{MoE}(h) = \sum_{i\in \text{Top-}k} g_i(h) \, \mathrm{FFN}_i(h),
\]

where \(g_i(h)\) is the gating weight assigned by the Router. By activating only a few experts per token, the effective FLOPs per token can be reduced from \(\mathcal{O}(E \cdot d \cdot d_\text{ff})\) to \(\mathcal{O}(k \cdot d \cdot d_\text{ff})\).

The Router is a lightweight module that predicts which experts should process a given token:
\[
g(h) = \mathrm{softmax}(h W_r),\qquad g \in \mathbb{R}^{E}.
\]
It selects the top-\(k\) experts according to the largest \(g_i\) values. The Router can also include auxiliary losses, such as load-balancing or importance losses, to encourage uniform expert utilization and avoid stragglers, which would increase memory or energy spikes.

By increasing the total number of experts \(E\) without increasing \(k\), it is possible to scale the model’s representational capacity while incurring only a small incremental energy cost per token. In practice, expert computations for different tokens are often batched across devices, but load imbalance can increase memory movement and create temporary energy spikes due to which careful load-balancing and token assignment become necessary to maintain efficiency.

%% file: Tables/Encoder-only-model-info.tex
\begin{table*}[h!]
\resizebox{\textwidth}{!}{%
\renewcommand{\arraystretch}{1.15}
\begin{tabular}{|l|c|c|c|c|c|l|}
\hline
\multicolumn{1}{|c|}{\textbf{Model}} & \textbf{\begin{tabular}[c]{@{}c@{}}\# of \\ Layers\end{tabular}} & \textbf{\begin{tabular}[c]{@{}c@{}}Hidden\\ Dimension\end{tabular}} & \textbf{\begin{tabular}[c]{@{}c@{}}Attention\\  Heads\end{tabular}} & \textbf{\begin{tabular}[c]{@{}c@{}}Feed-Forward \\ Dimension\end{tabular}} & \textbf{Parameters} & \multicolumn{1}{c|}{\textbf{Special Features}}                                                                                          \\ \hline
google-bert/bert-base-uncased        & 12 & 768  & 12 & 3072 & 110M & \begin{tabular}[c]{@{}l@{}}Uses {[}CLS{]} token for \\ classification\end{tabular} \\ \hline
google-bert/bert-large-uncased       & 24 & 1024 & 16 & 4096 & 340M & \begin{tabular}[c]{@{}l@{}}Larger variant of BERT with \\ higher representational capacity\end{tabular} \\ \hline
albert/albert-base-v2                & 12 & 768  & 12 & 3072 & 12M  & \begin{tabular}[c]{@{}l@{}}Parameter-sharing across \\ layers and factorized embedding\end{tabular} \\ \hline
albert/albert-large-v2               & 24 & 1024 & 16 & 4096 & 18M  & \begin{tabular}[c]{@{}l@{}}Deeper network with the same \\ parameter-sharing strategy\end{tabular} \\ \hline
distilbert/distilbert-base-uncased   & 6  & 768  & 12 & 3072 & 66M  & \begin{tabular}[c]{@{}l@{}}Distilled version of BERT with \\ 40\% fewer parameters\end{tabular} \\ \hline
distilroberta/distilroberta-base     & 6  & 768  & 12 & 3072 & 82M  & \begin{tabular}[c]{@{}l@{}}Distilled version of RoBERTa \\ retaining most performance\end{tabular} \\ \hline
FacebookAI/roberta-base              & 12 & 768  & 12 & 3072 & 125M & \begin{tabular}[c]{@{}l@{}}Improved pretraining and \\ dynamic masking\end{tabular} \\ \hline
FacebookAI/roberta-large             & 24 & 1024 & 16 & 4096 & 355M & \begin{tabular}[c]{@{}l@{}}Larger RoBERTa model with \\ improved pretraining\end{tabular} \\ \hline
\end{tabular}
}
\caption{Detailed architectural comparison of eight Encoder-Only models across key parameters like hidden dimension size, number of layers, number of parameters, etc.}
\label{fig:Encoder-only-model-details}
\end{table*}

%% file: Tables/Decoder-only-model-info.tex
\begin{table*}[h!]
\resizebox{\textwidth}{!}{%
\renewcommand{\arraystretch}{1.15}
\begin{tabular}{|l|p{3.5cm}|p{3.5cm}|p{3.5cm}|p{3.5cm}|}
\hline
\textbf{Architecture Detail} 
& \textbf{Qwen2.5-3B-Instruct} 
& \textbf{Phi-4-Mini-Instruct} 
& \textbf{Llama-3.2-3B-Instruct} 
& \textbf{Gemma-3-4B-IT} \\
\hline
\textbf{Parameters} 
& 3.09B total (2.77B non-embedding) 
& 4B 
& 3.21B 
& 4B \\
\hline
\textbf{Layers} 
& 36  
& 32 
& 28 
& 34  \\
\hline
\textbf{Hidden Size / Head Dim} 
& 2048 hidden, 128 per head 
& 3072 hidden, 128 per head 
& 3072 hidden, 128 per head 
& 2560 hidden, 128 per head\\
\hline
\textbf{Attention Structure} 
& GQA: 16 query heads, 2 KV heads; RoPE; QKV bias; output proj. biasless 
& GQA: 24 query heads, 8 KV heads; Fractional RoPE (25\% pos-agnostic); KV cache optimized 
& GQA: 24 query heads, 8 KV heads; RoPE; no bias in projections 
& Local+Global attention mix; Q-proj: 2048-d, K/V-proj: 1024-d; q\_norm, k\_norm applied \\
\hline
\textbf{MLP / FFN Dimension} 
& SwiGLU, 11008 (up+gate), 2048 down 
& SiLU, 16384 (gate+up), 8192 down 
& SiLU, 8192 up, 3072 down 
& GELU-Tanh, 10240 up, 2560 down \\
\hline
\textbf{Normalization} 
& RMSNorm, $\epsilon=1\mathrm{e}{-6}$, applied input + post-attn 
& RMSNorm, $\epsilon=1\mathrm{e}{-5}$, input + post-attn 
& RMSNorm, $\epsilon=1\mathrm{e}{-5}$, input + post-attn 
& RMSNorm, $\epsilon=1\mathrm{e}{-6}$, input + post-attn + pre/post-FFN \\
\hline
\textbf{Embeddings} 
& 151,936 vocab, 2048-d, tied in/out 
& 200,064 vocab, 3072-d, tied in/out (padding idx=199999) 
& 128,256 vocab, 3072-d, tied in/out 
& 262,208 vocab, 2560-d, tied in/out, scaled embeddings \\
\hline
\textbf{Context Length} 
& 32K tokens (gen up to 8K) 
& Long-context via KV optimization, tested up to $\sim$128K 
& 128K tokens, efficient GQA 
& 128K tokens; local layers span 1K, 1 global layer every 5 locals \\
\hline
\textbf{Special Features} 
& RoPE, SwiGLU, QKV bias, high multilingual coverage 
& GQA w/ reduced KV cache, fractional RoPE, tuned LR schedule 
& Optimized transformer, SFT+RLHF alignment, multilingual 
& Local-global hybrid attention, multimodal (SigLIP image encoder), Pan \& Scan for variable resolution \\
\hline
\end{tabular}%
}
\caption{Detailed architectural comparison of Decoder-Only Qwen2.5-3B, Phi-4-Mini, Llama-3.2-3B, and Gemma-3-4B instruction-tuned models.}
\end{table*}

%% file: Tables/Decoder-5000.tex
% \usepackage{multirow}
\begin{table*}[h!]
\resizebox{\textwidth}{!}{%
\renewcommand{\arraystretch}{1.5}
\begin{tabular}{|l|cc|cc|cc|cc|cc|cc|}
\hline
\multicolumn{1}{|c|}{Components} & \multicolumn{2}{c|}{Qwen2.5-3B fp32}              & \multicolumn{2}{c|}{Qwen2.5-3B fp16}              & \multicolumn{2}{c|}{Llama-3.2-3B fp32}              & \multicolumn{2}{c|}{Llama-3.2-3B fp16}              & \multicolumn{2}{c|}{Gemma-3-4B fp32}                & \multicolumn{2}{c|}{Gemma-3-4B fp16}                \\ \cline{2-13} 
\multicolumn{1}{|c|}{}                            & \multicolumn{1}{c|}{Mean} & Std. Dev              & \multicolumn{1}{c|}{Mean} & Std. Dev              & \multicolumn{1}{c|}{Mean} & Std. Dev              & \multicolumn{1}{c|}{Mean} & Std. Dev              & \multicolumn{1}{c|}{Mean} & Std. Dev              & \multicolumn{1}{c|}{Mean} & Std. Dev              \\ \hline
MLP                                               & 113.71                    & 6.687                 & 48.5                      & 2.13                  & 127.24                    & 1.94                  & 54.55                     & 2.01                  & 129.47                    & 1.86                  & 60.07                     & 2.13                  \\
Attention                                         & 27.64                     & 1.79                  & 33.99                     & 5.96                  & 58.31                     & 3.01                  & 23.88                     & 1.42                  & 42.82                     & 2.11                  & 36.84                     & 1.85                  \\
Input Layer Norm                                  & 2.59                      & 0.24                  & 6.42                      & 1.02                  & 3.2                       & 0.25                  & 3.61                      & 0.41                  & 3.74                      & 0.45                  & 4.42                      & 0.43                  \\
Attention Layer Norm                              & 2.81                      & 0.33                  & 6.8                       & 0.97                  & 3.2                       & 0.32                  & 3.85                      & 0.41                  & 3.49                      & 0.33                  & 4.52                      & 0.36                  \\ \hline
Capture (Block)                                   & 146.75                    & -                     & 95.71                     & -                     & 191.95                    & -                     & 85.89                     & -                     & 186.54                    & -                     & 114.83                    & -                     \\
Block                                             & 150.03                    & 3.39                  & 96.19                     & 5.15                  & 192.26                    & 24.92                 & 91.85                     & 1.98                  & 187.54                    & 3.91                  & 126.63                    & 4.54                  \\
\%Capture (Block)                                 & 97.81                     & \multicolumn{1}{l|}{} & 99.50                     & \multicolumn{1}{l|}{} & 99.84                     & \multicolumn{1}{l|}{} & 93.51                     & \multicolumn{1}{l|}{} & 99.47                     & \multicolumn{1}{l|}{} & 90.68                     & \multicolumn{1}{l|}{} \\ \hline
Final Layer Norm                                  & 2.96                      & 0.41                  & 6.94                      & 0.97                  & 3.2                       & 0.27                  & 4.2                       & 0.58                  & 3.5                       & 0.33                  & 4.53                      & 0.32                  \\
Embedding                                         & 0.81                      & 0.26                  & 0.74                      & 0.06                  & 0.69                      & 0.23                  & 0.68                      & 0.24                  & 1.52                      & 0.29                  & 1.28                      & 0.27                  \\
LLM Head                                          & 459.66                    & 2.18                  & 214.29                    & 3.56                  & 602.02                    & 2.97                  & 374.92                    & 2.96                  & 1040.63                   & 7.25                  & 480.64                    & 8.72                  \\ \hline
Model                                             & 5864.51                   & -                     & 3684.81                   & -                     & 5989.19                   & -                     & 2951.6                    & -                     & 7422.01                   & -                     & 4791.87                   & -                     \\
Capture (Model)                                   & 5995.27                   & 26.77                 & 3685.95                   & 29.11                 & 6029.32                   & 10.68                 & 3261.5                    & 30.99                 & 8086.96                   & 25.25                 & 5248.99                   & 89.34                 \\
\%Capture (Model)                                 & 97.82                     & \multicolumn{1}{l|}{} & 99.97                     & \multicolumn{1}{l|}{} & 99.33                     & \multicolumn{1}{l|}{} & 90.50                     & \multicolumn{1}{l|}{} & 91.78                     & \multicolumn{1}{l|}{} & 91.29                     & \multicolumn{1}{l|}{} \\ \hline
\end{tabular}
}
\caption{CLEAR demonstrating similar performance on RTX 5000 GPU for Decoder-only models(Qwen2.5-3B, Llama-3.2-3B, Gemma-3-4B) across fp16 and fp32 floating point precisions. The average Energy values and standard deviation values are in millijoules (mJ).}
\label{tab:decoder_5000_GPU}
\end{table*}

%% file: Appendix/Hardware.tex
\section{Hardware Specification \label{app:hardware}}

The experiments in this paper were carried out using NVIDIA’s Ada-Lovelace architecture GPUs, namely the RTX~5000~Ada and RTX~6000~Ada, in order to assess compute and energy performance. The Ada Lovelace architecture is fabricated on a custom 4\,nm TSMC process and includes third-generation RT cores and fourth-generation Tensor cores, enabling mixed precision operations (including FP8 with sparsity) that are integral to efficient transformer inference \cite{NVIDIA-RTX6000AdaDatasheet}. According to the official datasheets, the RTX~5000~Ada has 12{,}800 CUDA cores, 100 RT cores, 400 Tensor cores, 32\,GB of ECC GDDR6 memory over a 256-bit interface (providing $\sim$576\,GB/s bandwidth), and a total board power of approximately 250\,W \cite{PNY-RTX5000AdaDatasheet}. The RTX~6000~Ada model offers 18{,}176 CUDA cores, 142 RT cores, 568 Tensor cores, 48\,GB of ECC GDDR6 memory on a 384-bit interface ($\sim$960\,GB/s bandwidth), and has a board power of around 300\,W \cite{NVIDIA-RTX6000AdaDatasheet}. These hardware choices directly influence both the sustained compute throughput and the energy-per-FLOP metrics reported in our results. 

NVIDIA does not publish the precise NVML power sampling interval for the RTX~5000~Ada or RTX~6000~Ada. Prior work has shown that on modern NVIDIA GPUs, NVML’s power readouts are typically updated at a frequency of 20--50\,Hz (i.e., every 20--50\,ms), which constrains the granularity of fine-grained energy attribution \cite{Yang_2024}, \cite{nik2025energyconsciousllmdecodingimpact}.

\subsection{Nvidia RTX 5000 and RTX 6000}
We validate our methodology across three models on the NVIDIA RTX 5000 ADA GPU and observe a \%Capture exceeding 90\%, with minimal standard deviation across both fp16 and fp32 precisions. Interestingly, the energy consumption of normalization blocks remains higher for fp16 compared to fp32, similar to the trend observed on the NVIDIA RTX 6000.
Refer Tables \ref{tab:decoder_5000_GPU}, \ref{decoder-fp16-6000-tok}, \ref{decoder-fp16-6000-8tok}, \ref{decoder-fp32-6000-8tok}

%% file: Appendix/StdDev.tex
\begin{figure}[]
    \centering
    \includegraphics[width=\columnwidth]{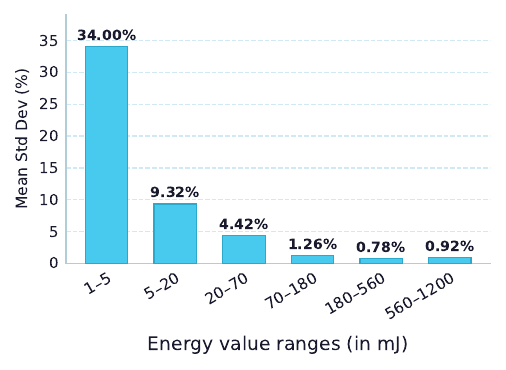}
    \caption{Standard Deviation across different Energy measurements. We observe a decrease in Standard Deviation in our Energy Measurement Approach with increasing Energy Values.}
    \label{fig:stddev Variation}
\end{figure}

\section{Variation in Standard Deviation}
\label{app:StdDev}
As shown in Fig.~\ref{fig:stddev Variation}, the standard deviation of energy measurements exhibits a higher relative deviation at lower energy values (around 1 mJ), primarily due to the limited precision of the NVML energy sensor. For measurements above 5 mJ, the deviation stabilizes to an acceptable range of approximately 9\%, and further decreases below 5\% for energies exceeding 20 mJ. This behavior arises because fixed sensor resolution introduces proportionally larger errors at smaller measurement scales.

%% file: Appendix/MultiToken.tex
\section{Multi Token \& KV Cache}
\label{app:MultiToken}
Using \textbf{\textcolor{CLEAR}{CLEAR}}, we extend our evaluation beyond the single-token \textit{Prefill} stage to study multi-token generation in decoder-only transformer models. Unlike single-token experiments, multi-token generation involves both the \textit{Prefill} and \textit{Decode} stages and introduces additional computational considerations due to Key–Value (KV) cache \cite{vaswani2023attentionneed} optimizations.

In autoregressive decoder models, tokens are generated sequentially. For each newly generated token, the attention mechanism must attend to all previously processed tokens in the sequence. In a naive implementation without caching, this requires recomputing the Key ($K$) and Value ($V$) projections for the entire sequence at every generation step. Consequently, both computational cost and energy consumption increase rapidly with sequence length due to repeated projection and attention operations.

\input{Appendix/Tables/MultiToken}

KV caching eliminates this redundancy by storing previously computed $K$ and $V$ tensors during the Prefill stage and reusing them during subsequent Decode steps. During decoding, only the Query ($Q$) corresponding to the newly generated token needs to be computed, while all previously computed keys and values are retrieved directly from the cache. This optimization significantly reduces redundant computation and memory traffic.

Our measurements show that multi-token generation exhibits distinct scaling behavior across the Prefill stage, Decode stage, and the full generation pipeline. The Prefill stage processes the entire input sequence and therefore shows a near-linear increase in both FLOPs and energy consumption as the sequence length grows. This behavior arises because fresh $Q$, $K$, and $V$ projections must be computed for every token in the sequence.

In contrast, the Decode stage remains nearly invariant to sequence length when KV caching is enabled. As shown in Table~\ref{tab: KV Cache}, the FLOPs remain constant at approximately 0.019 GFLOPs and the energy consumption remains nearly stable (approximately 29–33 mJ) across input lengths ranging from 1 to 512 tokens. This occurs because only the newly generated token is processed while all previous keys and values are reused from the cache.

When KV caching is disabled, however, the attention mechanism must recompute the entire set of $Q$, $K$, and $V$ projections at every step. As a result, the computational cost increases rapidly with sequence length. For example, as the input sequence increases from 1 to 512 tokens, FLOPs grow from 0.019 GFLOPs to 9.667 GFLOPs, while energy consumption increases from 28.05 mJ to 87.36 mJ.

\input{Appendix/Tables/MultiToken2}
These results demonstrate that KV caching substantially reduces redundant computation in the attention mechanism and leads to significant energy savings. At longer sequence lengths, KV cache achieves energy reductions exceeding 60\% compared to the uncached implementation. Figure~\ref{fig:multitoken} further illustrates this trend, showing both the percentage reduction in energy due to KV caching and the absolute difference in energy consumption with and without caching.

Overall, these observations highlight KV caching as a critical optimization for enabling efficient long-context autoregressive decoding. While the Prefill stage and portions of the output generation pipeline continue to scale with sequence length, KV caching ensures that the Decode stage remains computationally and energetically efficient.

%% file: Appendix/Tables/MultiToken.tex
% \usepackage{multirow}
\begin{table}[]
\resizebox{\columnwidth}{!}{%
\renewcommand{\arraystretch}{1.3}
{
\begin{tabular}{|c|ccc|ccc|}
\hline
\multirow{2}{*}{\textbf{\begin{tabular}[c]{@{}c@{}}Input\\  Token\\  Length\end{tabular}}} & \multicolumn{3}{c|}{\textbf{With KV Cache}}                                                                                                                                                   & \multicolumn{3}{c|}{\textbf{Without KV Cache}}                                                                                                                                               \\ \cline{2-7} 
                                                                                           & \multicolumn{1}{c|}{\textbf{\begin{tabular}[c]{@{}c@{}}Avg Energy\\ (mJ)\end{tabular}}} & \multicolumn{1}{c|}{\textbf{\begin{tabular}[c]{@{}c@{}}Std \\ Dev.\end{tabular}}} & \textbf{GFLOPS} & \multicolumn{1}{c|}{\textbf{\begin{tabular}[c]{@{}c@{}}Avg Energy\\ (mJ)\end{tabular}}} & \multicolumn{1}{c|}{\textbf{\begin{tabular}[c]{@{}c@{}}Std\\ Dev.\end{tabular}}} & \textbf{GFLOPS} \\ \hline
1                                                                                          & \multicolumn{1}{c|}{29.118}                                                             & \multicolumn{1}{c|}{0.515}                                                        & 0.019           & \multicolumn{1}{c|}{28.052}                                                             & \multicolumn{1}{c|}{0.474}                                                       & 0.019           \\
8                                                                                          & \multicolumn{1}{c|}{31.217}                                                             & \multicolumn{1}{c|}{0.524}                                                        & 0.019           & \multicolumn{1}{c|}{29.622}                                                             & \multicolumn{1}{c|}{0.682}                                                       & 0.151           \\
32                                                                                         & \multicolumn{1}{c|}{31.850}                                                             & \multicolumn{1}{c|}{0.511}                                                        & 0.019           & \multicolumn{1}{c|}{32.184}                                                             & \multicolumn{1}{c|}{0.580}                                                       & 0.604           \\
64                                                                                         & \multicolumn{1}{c|}{31.995}                                                             & \multicolumn{1}{c|}{0.513}                                                        & 0.019           & \multicolumn{1}{c|}{35.082}                                                             & \multicolumn{1}{c|}{0.690}                                                       & 1.208           \\
128                                                                                        & \multicolumn{1}{c|}{32.532}                                                             & \multicolumn{1}{c|}{0.239}                                                        & 0.019           & \multicolumn{1}{c|}{40.148}                                                             & \multicolumn{1}{c|}{0.696}                                                       & 2.417           \\
256                                                                                        & \multicolumn{1}{c|}{33.124}                                                             & \multicolumn{1}{c|}{0.785}                                                        & 0.019           & \multicolumn{1}{c|}{62.554}                                                             & \multicolumn{1}{c|}{1.591}                                                       & 4.834           \\
512                                                                                        & \multicolumn{1}{c|}{33.558}                                                             & \multicolumn{1}{c|}{1.335}                                                        & 0.019           & \multicolumn{1}{c|}{87.369}                                                             & \multicolumn{1}{c|}{2.220}                                                       & 9.667           \\ \hline
\end{tabular}
}}
\caption{Energy consumption and FLOP requirements for the attention mechanism in the Qwen2.5-3B model with and without KV cache across varying input sequence lengths. When KV cache is enabled, both energy usage and computational cost remain nearly constant, whereas disabling the KV cache leads to a sharp increase in FLOPs and energy as sequence length grows. The average Energy values and standard deviation values are in millijoules (mJ).}

\label{tab: KV Cache}
\end{table}

%% file: Appendix/Tables/MultiToken2.tex
% \usepackage{multirow}
\begin{table}[]
\resizebox{\columnwidth}{!}{%
\renewcommand{\arraystretch}{1.3}
{
\begin{tabular}{|c|cc|cc|cc|}
\hline
\multirow{2}{*}{\textbf{\begin{tabular}[c]{@{}c@{}}Input \\ Token \\ Length\end{tabular}}} & \multicolumn{2}{c|}{\textbf{Prefill}}                                                                      & \multicolumn{2}{c|}{\textbf{Next token Decode Stage}}                                                      & \multicolumn{2}{c|}{\textbf{Output Generation}}                                                            \\ \cline{2-7} 
                                                                                           & \multicolumn{1}{c|}{\textbf{\begin{tabular}[c]{@{}c@{}}Avg Energy \\ (mJ)\end{tabular}}} & \textbf{GFLOPs} & \multicolumn{1}{c|}{\textbf{\begin{tabular}[c]{@{}c@{}}Avg Energy \\ (mJ)\end{tabular}}} & \textbf{GFLOPs} & \multicolumn{1}{c|}{\textbf{\begin{tabular}[c]{@{}c@{}}Avg Energy \\ (mJ)\end{tabular}}} & \textbf{GFLOPs} \\ \hline
1                                                                                          & \multicolumn{1}{c|}{4329.42}                                                             & 6.17            & \multicolumn{1}{c|}{4472.20}                                                             & 6.17            & 4278.49                                                                                  & 6.17            \\
8                                                                                          & \multicolumn{1}{c|}{4531.80}                                                             & 49.38           & \multicolumn{1}{c|}{4543.73}                                                             & 6.17            & \multicolumn{1}{c|}{4517.82}                                                             & 45.02           \\
32                                                                                         & \multicolumn{1}{c|}{4829.55}                                                             & 197.51          & \multicolumn{1}{c|}{4600.39}                                                             & 6.17            & \multicolumn{1}{c|}{4749.55}                                                             & 178.22          \\
64                                                                                         & \multicolumn{1}{c|}{5273.63}                                                             & 395.03          & \multicolumn{1}{c|}{4675.02}                                                             & 6.17            & \multicolumn{1}{c|}{5254.10}                                                             & 355.82          \\
128                                                                                        & \multicolumn{1}{c|}{6964.34}                                                             & 790.06          & \multicolumn{1}{c|}{4749.80}                                                             & 6.17            & \multicolumn{1}{c|}{6895.61}                                                             & 711.02          \\
256                                                                                        & \multicolumn{1}{c|}{8959.38}                                                             & 1580.12         & \multicolumn{1}{c|}{4839.23}                                                             & 6.17            & \multicolumn{1}{c|}{8330.42}                                                             & 1421.42         \\
512                                                                                        & \multicolumn{1}{c|}{13940.02}                                                            & 3160.24         & \multicolumn{1}{c|}{5025.20}                                                             & 6.17            & \multicolumn{1}{c|}{13730.38}                                                            & 2842.23         \\ \hline
\end{tabular}
}}
\caption{Energy consumption and FLOP analysis of the Qwen2.5-3B model across varying input sequence lengths. We report measurements for three important phenomena: Prefill, Next-Token Decode (KV Cache Enabled) and Output Generation pipeline with 1 new token. The results show that while the FLOPs of the decode step remain constant due to KV reuse, its energy consumption increases only marginally with longer input lengths, in contrast to the Prefill and Output Generation where energy consumption scales rapidly with the input sequence length. The average Energy values  are in millijoules (mJ).}
\label{tab:Multi-token Decoder}
\end{table}

\arrayrulecolor{black} \color{black}

%% file: Appendix/Variants.tex
\section{Attention Variants and Optimizations}
\label{app:Variants}

\input{Appendix/Tables/VariantsOptim}

Using \textbf{\textcolor{CLEAR}{CLEAR}}, we compare three Attention implementations: \textit{Eager Attention}, \textit{Scaled Dot-Product Attention (SDPA)}, and \textit{Flash Attention}. Although all variants compute the same mathematical operation defined by Equation~\ref{eq:attention}, they differ significantly in how the computation is scheduled and executed on GPU hardware. These differences in execution strategy lead to measurable variations in memory access patterns, kernel launch behavior, and overall energy consumption.

\subsection*{Energy Comparison Across Attention Variants}

Across both models and all sequence lengths, Eager Attention consistently consumes the highest amount of energy. For example, in the Qwen2.5-3B model with 64 input tokens, Eager Attention consumes 45.4 mJ compared to 38.6 mJ for Flash Attention and 35.1 mJ for SDPA. A similar trend appears at larger sequence lengths: at 256 tokens, Eager Attention requires 70.4 mJ, while Flash Attention consumes 56.7 mJ and SDPA consumes 62.2 mJ.

The same behavior is observed in the Gemma3-4B model. At 64 tokens, Eager Attention consumes 67.0 mJ compared to 59.4 mJ for Flash Attention and 57.0 mJ for SDPA. At 256 tokens, Eager Attention increases to 100.2 mJ, while Flash Attention consumes 85.8 mJ and SDPA consumes 94.0 mJ.

This consistent gap in energy consumption is primarily caused by differences in memory usage and kernel execution. Eager Attention executes multiple independent GPU kernels for computing the query-key multiplication, softmax normalization, and value projection. Each of these operations launches separate GPU kernels and requires intermediate memory storage. As a result, the full attention matrix must be materialized in memory, which increases both memory traffic and kernel launch overhead.

SDPA improves upon this design by fusing some of the intermediate operations, which reduces kernel launch overhead. However, it still materializes the full attention matrix in GPU memory. Consequently, although SDPA is more efficient than Eager Attention, its memory traffic remains relatively high.

Flash Attention provides further improvements by reorganizing the computation into tiled blocks that fit into on-chip GPU memory. Instead of constructing the full attention matrix, the algorithm computes attention scores in smaller blocks and accumulates partial results. This approach significantly reduces memory movement between global memory and GPU cores. Since modern GPUs often spend more energy on memory access than on arithmetic computation, reducing memory traffic leads to lower overall energy consumption.

\subsection*{Impact of Sequence Length on Energy Efficiency}

Table~\ref{Attention-Optimization} also shows that total energy consumption increases as sequence length grows, but the \textit{energy per GFLOP} decreases. For example, in the Qwen2.5-3B model using Flash Attention, total energy increases from 38.6 mJ at 64 tokens to 56.7 mJ at 256 tokens. However, the energy per GFLOP decreases from 31.95 mJ/GFLOP to 11.72 mJ/GFLOP.

A similar trend can be observed in the Gemma3-4B model. Flash Attention consumes 59.4 mJ at 64 tokens and 85.8 mJ at 256 tokens, while the energy per GFLOP decreases from 29.50 mJ/GFLOP to 10.65 mJ/GFLOP.

The behavior can be explained by the energy modelling as a function of FLOPs described in Equation ~\ref{eq:model}. The total energy consumed by a component includes a fixed overhead term $E_0$ and a variable compute-dependent term. The fixed component includes kernel launch overhead, memory allocation, and GPU setup costs. When the sequence length is small, the fixed cost represents a large fraction of the total energy. As the sequence length increases, the computation grows faster than the fixed overhead, causing the energy cost to amortize across more operations. As a result, the energy consumed per unit of computation (mJ/GFLOP) decreases.

\subsection*{Effect of Torch Compile}

We isolate the effect of \textbf{torch.compile()}, which performs graph-level optimization and kernel fusion. Across both models, torch.compile() significantly reduces energy consumption for both Flash and Eager Attention.

For example, in Qwen2.5-3B with 256 tokens, Flash Attention consumes 56.7 mJ in the baseline implementation but only 47.6 mJ when torch.compile is enabled. Similarly, Eager Attention decreases from 70.4 mJ to 59.9 mJ after compilation. In the Gemma3-4B model at 256 tokens, Flash Attention drops from 85.8 mJ to 58.8 mJ, while Eager Attention drops from 100.2 mJ to 60.9 mJ.

The improvement occurs because torch.compile() analyzes the execution graph and merges multiple small GPU kernels into fewer, larger kernels. Larger kernels reduce kernel launch overhead and allow the GPU to maintain higher utilization. Kernel fusion also eliminates many intermediate memory reads and writes, which reduces memory traffic and therefore lowers energy consumption.

\subsection*{Effect of Precision}

The results also show that BF16 precision slightly increases energy consumption compared to FP16. For example, in Qwen2.5-3B with 64 tokens, BF16 Flash Attention consumes 48.3 mJ compared to 38.6 mJ for FP16 Flash Attention. In Gemma3-4B with 256 tokens, BF16 Flash Attention consumes 101.6 mJ compared to 85.8 mJ for FP16 Flash Attention.

This difference arises because some GPU architectures internally convert BF16 operations to FP32 during execution. These conversion steps introduce additional arithmetic operations and memory movement, which slightly increases the total energy consumed.

\subsection*{Aggressive Kernel Optimization}

The most significant reduction in energy consumption occurs with the \textbf{Max Autotune} and \textbf{Reduced Overhead} optimization modes. These configurations produce the lowest energy values across all models and sequence lengths.

For example, in Qwen2.5-3B with 256 tokens, Max Autotune reduces energy consumption to 33.5 mJ and Reduced Overhead reduces it to 36.6 mJ. Similarly, in Gemma3-4B with 256 tokens, energy decreases to 41.6 mJ with Max Autotune and 44.7 mJ with Reduced Overhead. These values are significantly lower than the baseline implementations.

These optimizations work by aggressively fusing operations and selecting hardware-specific kernel implementations. Max Autotune performs extensive kernel benchmarking during compilation to determine the most efficient kernel configurations for the current hardware. This includes searching over tile sizes, block sizes, memory layouts, and execution schedules. Reduced Overhead removes profiling and synchronization steps that are typically inserted during model execution, which further reduces kernel launch overhead.

\subsection*{Limitations of FLOP Measurement}

For the Max Autotune and Reduced Overhead configurations, it is not possible to reliably measure FLOPs using standard profiling tools. These optimizations generate fused kernels that combine multiple operations such as matrix multiplication, softmax normalization, and scaling into a single hardware-level kernel. Because these fused kernels no longer correspond to individual high-level operators, existing profiling tools cannot accurately attribute a FLOP count to each operation. As a result, the FLOP values for these configurations are not reported in Table~\ref{Attention-Optimization}.
\color{black}

%% file: Appendix/Tables/VariantsOptim.tex
% \usepackage{multirow}

\begin{table*}[]
\resizebox{\textwidth}{!}{%
\renewcommand{\arraystretch}{1.2}
{
{
\begin{tabular}{|l|l|c|c|c|c|}
\hline
\begin{tabular}[c]{@{}l@{}}Model \& \\ Input Length\end{tabular}                             & \multicolumn{1}{c|}{Attention Setup} & \begin{tabular}[c]{@{}c@{}}Energy Consumption\\ (mJ)\end{tabular} & \begin{tabular}[c]{@{}c@{}}Std. \\ Deviation\end{tabular} & \begin{tabular}[c]{@{}c@{}}FLOPs\\ (in GFLOPs)\end{tabular} & \textbf{Energy (mJ) / GFLOP} \\ \hline
\multirow{8}{*}{\begin{tabular}[c]{@{}l@{}}Qwen2.5 - 3B  \\ 64 input tokens\end{tabular}}   & FP16 Flash Attention                 & 38.609                                                            & 0.348                                                     & 1.208                                                       & 31.950                       \\
                                                                                            & FP16 Eager Attention                 & 45.398                                                            & 0.564                                                     & 1.242                                                       & 36.552                       \\
                                                                                            & FP16 SDPA                            & 35.061                                                            & 0.378                                                     & 1.208                                                       & 29.014                       \\
                                                                                            & FP16 Flash with torch.compile()      & 29.100                                                            & 0.320                                                     & 1.208                                                       & 24.090                       \\
                                                                                            & FP16 Eager with torch.compile()      & 31.675                                                            & 0.163                                                     & 1.242                                                       & 25.513                       \\
                                                                                            & BF16 Flash Attention                 & 48.289                                                            & 1.247                                                     & 2.417                                                       & 19.980                       \\
                                                                                            & FP16 Eager Max Autotune              & 19.679                                                            & 0.664                                                     & -                                                           & -                            \\
                                                                                            & F16 Eager Reduce Overhead            & 19.640                                                            & 0.765                                                     & -                                                           & -                            \\ \hline
\multirow{8}{*}{\begin{tabular}[c]{@{}l@{}}Qwen2.5 - 3B  \\ 128 input tokens\end{tabular}}  & FP16 Flash Attention                 & 44.169                                                            & 0.764                                                     & 2.417                                                       & 18.276                       \\
                                                                                            & FP16 Eager Attention                 & 53.766                                                            & 1.506                                                     & 2.551                                                       & 21.074                       \\
                                                                                            & FP16 SDPA                            & 39.577                                                            & 0.642                                                     & 2.417                                                       & 16.376                       \\
                                                                                            & FP16 Flash with torch.compile()      & 33.337                                                            & 0.441                                                     & 2.416                                                       & 13.799                       \\
                                                                                            & FP16 Eager with torch.compile()      & 38.881                                                            & 0.947                                                     & 2.550                                                       & 15.247                       \\
                                                                                            & BF16 Flash Attention                 & 48.289                                                            & 1.247                                                     & 2.417                                                       & 19.980                       \\
                                                                                            & FP16 Eager Max Autotune              & 24.977                                                            & 0.318                                                     & -                                                           & -                            \\
                                                                                            & F16 Eager Reduce Overhead            & 24.233                                                            & 0.881                                                     & -                                                           & -                            \\ \hline
\multirow{8}{*}{\begin{tabular}[c]{@{}l@{}}Qwen2.5 - 3B   \\ 256 input tokens\end{tabular}} & FP16 Flash Attention                 & 56.665                                                            & 1.131                                                     & 4.834                                                       & 11.723                       \\
                                                                                            & FP16 Eager Attention                 & 70.411                                                            & 0.585                                                     & 5.372                                                       & 13.108                       \\
                                                                                            & FP16 SDPA                            & 62.223                                                            & 1.319                                                     & 4.834                                                       & 12.873                       \\
                                                                                            & FP16 Flash with torch.compile()      & 47.558                                                            & 0.702                                                     & 4.832                                                       & 9.843                        \\
                                                                                            & FP16 Eager with torch.compile()      & 59.949                                                            & 0.818                                                     & 5.369                                                       & 11.166                       \\
                                                                                            & BF16 Flash Attention                 & 48.289                                                            & 1.247                                                     & 2.417                                                       & 19.980                       \\
                                                                                            & FP16 Eager Max Autotune              & 33.500                                                            & 0.874                                                     & -                                                           & -                            \\
                                                                                            & F16 Eager Reduce Overhead            & 36.626                                                            & 1.095                                                     & -                                                           & -                            \\ \hline
\multirow{8}{*}{\begin{tabular}[c]{@{}l@{}}Gemma3 - 4B   \\ 64 input tokens\end{tabular}}   & FP16 Flash Attention                 & 59.416                                                            & 0.882                                                     & 2.014                                                       & 29.498                       \\
                                                                                            & FP16 Eager Attention                 & 66.962                                                            & 0.694                                                     & 2.048                                                       & 32.699                       \\
                                                                                            & FP16 SDPA                            & 57.048                                                            & 1.035                                                     & 2.014                                                       & 28.322                       \\
                                                                                            & FP16 Flash with torch.compile()      & 32.953                                                            & 0.461                                                     & 2.013                                                       & 16.368                       \\
                                                                                            & FP16 Eager with torch.compile()      & 34.322                                                            & 0.383                                                     & 2.047                                                       & 16.769                       \\
                                                                                            & BF16 Flash Attention                 & 67.633                                                            & 0.986                                                     & 2.014                                                       & 33.577                       \\
                                                                                            & FP16 Eager Max Autotune              & 23.842                                                            & 0.762                                                     & -                                                           & -                            \\
                                                                                            & F16 Eager Reduce Overhead            & 23.827                                                            & 0.841                                                     & -                                                           & -                            \\ \hline
\multirow{8}{*}{\begin{tabular}[c]{@{}l@{}}Gemma3 - 4B   \\ 128 input tokens\end{tabular}}  & FP16 Flash Attention                 & 69.228                                                            & 0.749                                                     & 4.029                                                       & 17.184                       \\
                                                                                            & FP16 Eager Attention                 & 76.126                                                            & 0.658                                                     & 4.163                                                       & 18.287                       \\
                                                                                            & FP16 SDPA                            & 65.785                                                            & 0.504                                                     & 4.029                                                       & 16.330                       \\
                                                                                            & FP16 Flash with torch.compile()      & 43.885                                                            & 0.558                                                     & 4.027                                                       & 10.899                       \\
                                                                                            & FP16 Eager with torch.compile()      & 44.267                                                            & 0.562                                                     & 4.161                                                       & 10.639                       \\
                                                                                            & BF16 Flash Attention                 & 76.328                                                            & 1.326                                                     & 4.029                                                       & 18.947                       \\
                                                                                            & FP16 Eager Max Autotune              & 30.591                                                            & 0.852                                                     & -                                                           & -                            \\
                                                                                            & F16 Eager Reduce Overhead            & 31.651                                                            & 0.851                                                     & -                                                           & -                            \\ \hline
\multirow{8}{*}{\begin{tabular}[c]{@{}l@{}}Gemma3 - 4B   \\ 256 input tokens\end{tabular}}  & FP16 Flash Attention                 & 85.843                                                            & 1.854                                                     & 8.057                                                       & 10.654                       \\
                                                                                            & FP16 Eager Attention                 & 100.152                                                           & 2.179                                                     & 8.594                                                       & 11.653                       \\
                                                                                            & FP16 SDPA                            & 93.987                                                            & 1.188                                                     & 8.057                                                       & 11.665                       \\
                                                                                            & FP16 Flash with torch.compile()      & 58.833                                                            & 0.814                                                     & 8.053                                                       & 7.306                        \\
                                                                                            & FP16 Eager with torch.compile()      & 60.907                                                            & 0.696                                                     & 8.590                                                       & 7.091                        \\
                                                                                            & BF16 Flash Attention                 & 101.571                                                           & 3.304                                                     & 8.057                                                       & 12.607                       \\
                                                                                            & FP16 Eager Max Autotune              & 41.614                                                            & 0.619                                                     & -                                                           & -                            \\
                                                                                            & F16 Eager Reduce Overhead            & 44.674                                                            & 0.161                                                     & -                                                           & -                            \\ \hline
\end{tabular}
}}}
\caption{ {Average Energy Consumption, FLOPs and Energy(mJ)/GFLOPs ratio for Gemma3-4B and Qwen2.5-3B models across input token lengths of 64, 128 and 256. We demonstrate results for 3 Attention Variants (SDPA, Eager, Flash) along with Optimizations such as Torch Compile, Max Autotune and Reduced Overhead. The average Energy values and standard deviation values are in millijoules (mJ). }}
\label{Attention-Optimization}
\end{table*}

%% file: Appendix/Tables/Tables_appendix.tex
\section{Additional Results}
\label{app:Results}

\input{Appendix/Tables/Encoder_models_fp16}

\input{Appendix/Tables/Encoder_models_fp16_large}

\input{Appendix/Tables/Encoder_models_fp32}

\input{Appendix/Tables/Encoder_Models_fp32_large}

\input{Appendix/Tables/Encoder_models_token_length}

\input{Appendix/Tables/Encoder_models_token_length_large}

\input{Appendix/Tables/Decoder_fp16_6000}

\input{Appendix/Tables/Decoder_fp16_8token}
\input{Appendix/Tables/Decoder_fp32_8token}

%% file: Appendix/Tables/Encoder_models_fp16.tex
% \usepackage{multirow}
\begin{table*}[]
\resizebox{\textwidth}{!}{%
\renewcommand{\arraystretch}{1.4}
\begin{tabular}{|l|cc|cc|cc|cc|}
\hline
\multicolumn{1}{|c|}{\multirow{2}{*}{Components}} & \multicolumn{2}{c|}{ALBERT}          & \multicolumn{2}{c|}{BERT}            & \multicolumn{2}{c|}{DistilBERT}                    & \multicolumn{2}{c|}{RoBERTa}         \\ \cline{2-9} 
\multicolumn{1}{|c|}{}                            & \multicolumn{1}{c|}{Avg.} & Std. Dev & \multicolumn{1}{c|}{Avg.} & Std. Dev & \multicolumn{1}{c|}{Avg.} & Std. Dev               & \multicolumn{1}{c|}{Avg.} & Std. Dev \\ \hline
Attention Block                                   & 12.074                    & 0.280    & 13.676                    & 0.370    & 10.550                    & 0.369                  & 13.534                    & 0.319    \\
FFN (Intermediate)                                & 1.510                     & 0.441    & 2.765                     & 0.192    & \multirow{2}{*}{8.764}    & \multirow{2}{*}{0.442} & 2.565                     & 0.321    \\
FFN (Output)                                      & 1.848                     & 0.050    & 4.944                     & 0.343    &                           &                        & 4.875                     & 0.401    \\
Norm. (All)                                       & 7.317                     & 0.531    & 6.216                     & 0.845    & 7.135                     & 0.369                  & 6.352                     & 0.623    \\ \hline
Captured (Block)                                  & 22.748                    & -        & 27.601                    & -        & 26.449                    & -                      & 27.326                    & -        \\
Measured (Block)                                             & 30.627                    & 0.562    & 29.802                    & 0.607    & 28.551                    & 0.448                  & 29.797                    & 0.642    \\
\hline
\end{tabular}
}
\caption{Comparison of FP16 Performance across ALBERT, BERT, DistilBERT, RoBERTa. The average Energy values and standard deviation values are in millijoules (mJ).}
\end{table*}

%% file: Appendix/Tables/Encoder_models_fp16_large.tex
% \usepackage{multirow}
\begin{table*}[]
\resizebox{\textwidth}{!}{%
\renewcommand{\arraystretch}{1.4}
\begin{tabular}{|l|cc|cc|cc|cc|}
\hline
\multicolumn{1}{|c|}{\multirow{2}{*}{Components}} & \multicolumn{2}{c|}{ALBERT - Large}  & \multicolumn{2}{c|}{BERT - Large}    & \multicolumn{2}{c|}{Distil RoBERTa}                & \multicolumn{2}{c|}{RoBERTa - Large} \\ \cline{2-9} 
\multicolumn{1}{|c|}{}                            & \multicolumn{1}{c|}{Avg.} & Std. Dev & \multicolumn{1}{c|}{Avg.} & Std. Dev & \multicolumn{1}{c|}{Avg.} & Std. Dev               & \multicolumn{1}{c|}{Avg.} & Std. Dev \\ \hline
Attention Block                                   & 14.861                    & 0.423    & 16.903                    & 0.397    & 13.226                    & 0.349                  & 15.547                    & 0.370    \\
FFN (Intermediate)                                & 2.928                     & 0.743    & 4.877                     & 0.456    & \multirow{2}{*}{7.413}    & \multirow{2}{*}{0.396} & 4.249                     & 0.115    \\
FFN (Output)                                      & 3.939                     & 0.710    & 8.093                     & 0.462    &                           &                        & 7.741                     & 0.443    \\
Norm. (All)                                       & 5.814                     & 0.652    & 4.171                     & 0.872    & 4.274                     & 0.711                  & 4.721                     & 0.671    \\ \hline
Captured (Block)                                  & 27.542                    & -        & 34.044                    & -        & 24.913                    & -                      & 32.259                    & -        \\
Block                                             & 34.293                    & 0.493    & 34.590                    & 0.709    & 28.670                    & 0.511                  & 33.854                    & 0.833    \\
\hline
\end{tabular}
}
\caption{Comparison of FP16 Performance across variants- ALBERT Large, BERT Large, DistilRoBERTa, RoBERTa Large. The average Energy values and standard deviation values are in millijoules (mJ). }
\end{table*}

%% file: Appendix/Tables/Encoder_models_fp32.tex
% \usepackage{multirow}
\begin{table*}[]
\resizebox{\textwidth}{!}{%
\renewcommand{\arraystretch}{1.4}
\begin{tabular}{|l|cc|cc|cc|cc|}
\hline
\multicolumn{1}{|c|}{\multirow{2}{*}{Components}} & \multicolumn{2}{c|}{ALBERT}          & \multicolumn{2}{c|}{BERT}            & \multicolumn{2}{c|}{DistilBERT}                    & \multicolumn{2}{c|}{RoBERTa}         \\ \cline{2-9} 
\multicolumn{1}{|c|}{}                            & \multicolumn{1}{c|}{Avg.} & Std. Dev & \multicolumn{1}{c|}{Avg.} & Std. Dev & \multicolumn{1}{c|}{Avg.} & Std. Dev               & \multicolumn{1}{c|}{Avg.} & Std. Dev \\ \hline
Attention Block                                   & 18.573                    & 0.517    & 19.155                    & 0.436    & 17.772                    & 0.412                  & 17.870                    & 0.327    \\
FFN (Intermediate)                                & 4.969                     & 0.980    & 7.163                     & 0.880    & \multirow{2}{*}{17.486}   & \multirow{2}{*}{0.774} & 6.446                     & 0.724    \\
FFN (Output)                                      & 5.066                     & 0.935    & 9.528                     & 0.646    &                           &                        & 8.829                     & 0.620    \\
Norm. (All)                                       & 5.267                     & 0.432    & 4.023                     & 0.972    & 5.872                     & 0.861                  & 4.721                     & 0.671    \\ \hline
Captured (Block)                                  & 33.875                    & -        & 39.869                    & -        & 41.130                    & -                      & 37.867                    & -        \\
Block                                             & 40.282                    & 0.779    & 42.030                    & 0.881    & 46.199                    & 0.755                  & 40.054                    & 1.545    \\
\hline
\end{tabular}
}
\caption{Comparison of FP32 Performance across ALBERT, BERT, DistilBERT, RoBERTa. The average Energy values and standard deviation values are in millijoules (mJ).}
\end{table*}

%% file: Appendix/Tables/Encoder_Models_fp32_large.tex
% \usepackage{multirow}
\begin{table*}[]
\resizebox{\textwidth}{!}{%
\renewcommand{\arraystretch}{1.4}
\begin{tabular}{|l|cc|cc|cc|cc|}
\hline
\multicolumn{1}{|c|}{\multirow{2}{*}{Components}} & \multicolumn{2}{c|}{ALBERT - Large}  & \multicolumn{2}{c|}{BERT - Large}    & \multicolumn{2}{c|}{Distil RoBERTa}  & \multicolumn{2}{c|}{RoBERTa - Large} \\ \cline{2-9} 
\multicolumn{1}{|c|}{}                            & \multicolumn{1}{c|}{Avg.} & Std. Dev & \multicolumn{1}{c|}{Avg.} & Std. Dev & \multicolumn{1}{c|}{Avg.} & Std. Dev & \multicolumn{1}{c|}{Avg.} & Std. Dev \\ \hline
Attention Block                                   & 23.380                    & 0.545    & 25.777                    & 0.546    & 19.888                    & 0.493    & 24.569                    & 0.572    \\
FFN (Intermediate)                                & 8.139                     & 1.049    & 9.679                     & 0.929    & 7.275                     & 0.924    & 9.145                     & 0.949    \\
FFN (Output)                                      & 8.090                     & 0.843    & 12.952                    & 0.799    & 9.636                     & 0.679    & 12.124                    & 0.817    \\
Norm. (All)                                       & 4.479                     & 0.512    & 4.413                     & 0.872    & 3.761                     & 0.843    & 4.310                     & 0.871    \\ \hline
Captured (Block)                                  & 44.089                    & -        & 52.821                    & -        & 40.561                    & -        & 50.148                    & -        \\
Block                                             & 60.662                    & 0.918    & 58.006                    & 0.841    & 42.234                    & 1.687    & 55.608                    & 1.023    \\
\hline
\end{tabular}
}
\caption{Comparison of FP32 Performance across large variants of ALBERT, BERT, DistilBERT, RoBERTa. The average Energy values and standard deviation values are in millijoules (mJ).}
\end{table*}

%% file: Appendix/Tables/Encoder_models_token_length.tex
% \usepackage{multirow}
\begin{table*}[]
\resizebox{\textwidth}{!}{%
\renewcommand{\arraystretch}{1}
\begin{tabular}{|ccccccccc|}
\hline
\multicolumn{1}{|c|}{\multirow{2}{*}{Component}} & \multicolumn{2}{c|}{32 Tokens}                             & \multicolumn{2}{c|}{64 Tokens}           & \multicolumn{2}{c|}{96 Tokens}                             & \multicolumn{2}{c|}{128 Tokens}       \\ \cline{2-9} 
\multicolumn{1}{|c|}{}                           & \multicolumn{1}{c|}{Avg.} & \multicolumn{1}{c|}{Std. Dev.} & Avg.    & \multicolumn{1}{c|}{Std. Dev.} & \multicolumn{1}{c|}{Avg.} & \multicolumn{1}{c|}{Std. Dev.} & \multicolumn{1}{c|}{Avg.} & Std. Dev. \\ \hline
\multicolumn{9}{|c|}{\multirow{2}{*}{ALBERT-Base}}                                                                                                                                                                                                            \\
\multicolumn{9}{|c|}{}                                                                                                                                                                                                                                        \\ \hline
\multicolumn{1}{|c|}{Attention Block}            & 15.094                    & \multicolumn{1}{c|}{0.404}     & 15.930  & \multicolumn{1}{c|}{0.354}     & 18.582                    & \multicolumn{1}{c|}{0.319}     & 19.861                    & 0.468     \\
\multicolumn{1}{|c|}{FFN (Intermediate)}         & 2.306                     & \multicolumn{1}{c|}{0.448}     & 2.205   & \multicolumn{1}{c|}{0.475}     & 3.341                     & \multicolumn{1}{c|}{0.308}     & 3.650                     & 0.594     \\
\multicolumn{1}{|c|}{FFN (Output)}               & 2.361                     & \multicolumn{1}{c|}{0.423}     & 2.891   & \multicolumn{1}{c|}{0.599}     & 4.608                     & \multicolumn{1}{c|}{0.837}     & 4.162                     & 0.751     \\
\hline
\end{tabular}
}
\caption{Energy consumption trends with varying input token lengths tokens) for encoder only (Base) models. The average Energy values and standard deviation values are in millijoules (mJ).}
\end{table*}

%% file: Appendix/Tables/Encoder_models_token_length_large.tex
% \usepackage{multirow}
\begin{table*}[]
\resizebox{\textwidth}{!}{%
\renewcommand{\arraystretch}{1}
\begin{tabular}{|ccccccccc|}
\hline
\multicolumn{1}{|c|}{\multirow{2}{*}{Component}} & \multicolumn{2}{c|}{32 Tokens}                             & \multicolumn{2}{c|}{64 Tokens}           & \multicolumn{2}{c|}{96 Tokens}                             & \multicolumn{2}{c|}{128 Tokens}       \\ \cline{2-9} 
\multicolumn{1}{|c|}{}                           & \multicolumn{1}{c|}{Avg.} & \multicolumn{1}{c|}{Std. Dev.} & Avg.    & \multicolumn{1}{c|}{Std. Dev.} & \multicolumn{1}{c|}{Avg.} & \multicolumn{1}{c|}{Std. Dev.} & \multicolumn{1}{c|}{Avg.} & Std. Dev. \\ \hline
\multicolumn{9}{|c|}{\multirow{2}{*}{ALBERT-Large}}                                                                                                                                                                                                           \\
\multicolumn{9}{|c|}{}                                                                                                                                                                                                                                        \\ \hline
\multicolumn{1}{|c|}{Attention Block}            & 17.730                    & \multicolumn{1}{c|}{0.472}     & 19.696  & \multicolumn{1}{c|}{0.645}     & 22.541                    & \multicolumn{1}{c|}{0.529}     & 24.279                    & 0.435     \\
\multicolumn{1}{|c|}{FFN (Intermediate)}         & 3.364                     & \multicolumn{1}{c|}{0.812}     & 3.399   & \multicolumn{1}{c|}{0.927}     & 5.134                     & \multicolumn{1}{c|}{0.888}     & 4.352                     & 0.821     \\
\multicolumn{1}{|c|}{FFN (Output)}               & 4.117                     & \multicolumn{1}{c|}{0.688}     & 4.724   & \multicolumn{1}{c|}{0.900}     & 5.424                     & \multicolumn{1}{c|}{0.956}     & 6.064                     & 0.913     \\
\hline
\end{tabular}
}
\caption{Energy consumption trends with varying input token lengths tokens) for encoder only (Large) models. The average Energy values and standard deviation values are in millijoules (mJ).}
\end{table*}

%% file: Appendix/Tables/Decoder_fp16_6000.tex
% \usepackage{multirow}
\begin{table*}[]
\resizebox{\textwidth}{!}{%
\renewcommand{\arraystretch}{1}
\begin{tabular}{ccccccccc|}
\hline
\multicolumn{1}{|c|}{\multirow{2}{*}{Component}} & \multicolumn{2}{c|}{32 Tokens}            & \multicolumn{2}{c|}{64 Tokens}             & \multicolumn{2}{c|}{96 Tokens}             & \multicolumn{2}{c|}{128 Tokens}   \\ \cline{2-9} 
\multicolumn{1}{|c|}{}                           & Avg.     & \multicolumn{1}{c|}{Std. Dev.} & Avg.      & \multicolumn{1}{c|}{Std. Dev.} & Avg.      & \multicolumn{1}{c|}{Std. Dev.} & Avg.      & Std. Dev.             \\ \hline
\multicolumn{9}{|c|}{\multirow{2}{*}{Gemma-3  4B}}                                                                                                                                                                         \\
\multicolumn{9}{|c|}{}                                                                                                                                                                                                     \\ \hline
\multicolumn{1}{|c|}{Attention Block}            & 47.312   & \multicolumn{1}{c|}{1.421}     & 53.619    & \multicolumn{1}{c|}{1.672}     & 64.300    & \multicolumn{1}{c|}{1.381}     & 65.488    & 1.050                 \\
\multicolumn{1}{|c|}{MLP}                        & 73.218   & \multicolumn{1}{c|}{1.043}     & 75.700    & \multicolumn{1}{c|}{0.872}     & 81.383    & \multicolumn{1}{c|}{0.776}     & 83.230    & 0.995                 \\
\multicolumn{1}{|c|}{Norm. (All)}                & 26.144   & \multicolumn{1}{c|}{0.533}     & 27.853    & \multicolumn{1}{c|}{0.456}     & 30.892    & \multicolumn{1}{c|}{0.394}     & 32.878    & 0.877                 \\ \hline
\multicolumn{1}{|c|}{Captured (Block)}           & 146.674  & \multicolumn{1}{c|}{-}         & 157.173   & \multicolumn{1}{c|}{-}         & 176.575   & \multicolumn{1}{c|}{-}         & 181.596   & -                     \\
\hline

\end{tabular}}
\caption{Energy of Decoder Model Components using CLEAR on RTX 6000 GPU for models(Qwen2.5-3B, Llama-3.2-3B, Gemma-3-4B, Phi3-4B) with fp16 across token length. The average Energy values and standard deviation values are in millijoules (mJ).}
\label{decoder-fp16-6000-tok}

\end{table*}

%% file: Appendix/Tables/Decoder_fp16_8token.tex
% \usepackage{multirow}
\begin{table*}[]
\resizebox{\textwidth}{!}{%
\renewcommand{\arraystretch}{1.4}
\begin{tabular}{|c|cc|cc|cc|cc|}
\hline
\multirow{2}{*}{Components} & \multicolumn{2}{c|}{Qwen2.5} & \multicolumn{2}{c|}{Llama3.2} & \multicolumn{2}{c|}{Gemma 3} & \multicolumn{2}{c|}{Phi-3 4B} \\ \cline{2-9} 
                            & Avg.         & Std. Dev.     & Avg.          & Std. Dev.     & Avg.         & Std. Dev.     & Avg.          & Std. Dev.     \\ \hline
Attention Block             & 32.791       & 1.090         & 42.842        & 0.427         & 45.215       & 0.632         & 62.817        & 0.831         \\
MLP                         & 62.257       & 1.130         & 64.830        & 1.774         & 66.196       & 1.642         & 132.590       & 2.548         \\
Norm. (All)                 & 9.769        & 0.812         & 9.801         & 0.845         & 23.643       & 1.117         & 20.437        & 1.940         \\ \hline
Captured (Block)            & 104.818      & -             & 117.473       & -             & 135.054      & -             & 215.843       & -             \\
Block                       & 113.095      & 1.855         & 121.860       & 2.067         & 149.414      & 3.911         & 214.413       & 1.106         \\
\%Capture(Block)            & 92.682       & -             & 96.400        & -             & 90.389       & -             & 100.667       & -             \\
\hline
\end{tabular}}
\caption{Energy of Decoder Model Components using CLEAR on RTX 6000 GPU for models(Qwen2.5-3B, Llama-3.2-3B, Gemma-3-4B, Phi3-4B) with fp16 and 8 token length. The average Energy values and standard deviation values are in millijoules (mJ).}
\label{decoder-fp16-6000-8tok}

\end{table*}

%% file: Appendix/Tables/Decoder_fp32_8token.tex
% \usepackage{multirow}
\begin{table*}[]
\resizebox{\textwidth}{!}{%
\renewcommand{\arraystretch}{1.4}
\begin{tabular}{|c|cc|cc|cc|cc|}
\hline
\multirow{2}{*}{Components} & \multicolumn{2}{c|}{Qwen2.5} & \multicolumn{2}{c|}{Llama} & \multicolumn{2}{c|}{Gemma} & \multicolumn{2}{c|}{Phi} \\ \cline{2-9} 
                            & Avg.         & Std. Dev.     & Avg.        & Std. Dev.    & Avg.         & Std. Dev.   & Avg.       & Std. Dev.   \\ \hline
Attention Block             & 49.584       & 0.891         & 90.613      & 0.930        & 83.220       & 1.302       & 88.889     & 0.893       \\
MLP                         & 132.007      & 1.069         & 149.203     & 1.259        & 144.130      & 0.776       & 143.361    & 0.984       \\
Norm. (All)                 & 7.068        & 0.206         & 7.310       & 0.228        & 16.947       & 1.480       & 7.736      & 0.220       \\ \hline
Captured (Block)            & 188.659      & -             & 247.125     & -            & 244.297      & -           & 239.986    & -           \\
Block                       & 187.446      & 1.181         & 241.463     & 0.959        & 257.972      & 6.467       & 249.932    & 6.578       \\
\%Capture (Block)           & 100.647      & -             & 102.345     & -            & 94.699       & -           & 96.020     & -           \\
\hline
\end{tabular}}
\caption{Energy of Decoder Model Components using CLEAR on RTX 6000 GPU for models(Qwen2.5-3B, Llama-3.2-3B, Gemma-3-4B, Phi3-4B) with fp32 and 8 token length. The average Energy values and standard deviation values are in millijoules (mJ).}
\label{decoder-fp32-6000-8tok}

\end{table*}

%% file: custom.bib
@inproceedings{Alizadeh_2024, series={CAIN 2024},
   title={Green AI: a Preliminary Empirical Study on Energy Consumption in DL Models Across Different Runtime Infrastructures},
   url={http://dx.doi.org/10.1145/3644815.3644967},
   DOI={10.1145/3644815.3644967},
   booktitle={Proceedings of the IEEE/ACM 3rd International Conference on AI Engineering - Software Engineering for AI},
   publisher={ACM},
   author={Alizadeh, Negar and Castor, Fernando},
   year={2024},
   month=apr, pages={134–139},
   collection={CAIN 2024} }

@article{S_nchez_Momp__2025,
   title={Green MLOps to Green GenOps: An Empirical Study of Energy Consumption in Discriminative and Generative AI Operations},
   volume={16},
   ISSN={2078-2489},
   url={http://dx.doi.org/10.3390/info16040281},
   DOI={10.3390/info16040281},
   number={4},
   journal={Information},
   publisher={MDPI AG},
   author={Sánchez-Mompó, Adrián and Mavromatis, Ioannis and Li, Peizheng and Katsaros, Konstantinos and Khan, Aftab},
   year={2025},
   month=mar, pages={281} }

@misc{rajput2024enhancingenergyawarenessdeeplearning,
      title={Enhancing Energy-Awareness in Deep Learning through Fine-Grained Energy Measurement}, 
      author={Saurabhsingh Rajput and Tim Widmayer and Ziyuan Shang and Maria Kechagia and Federica Sarro and Tushar Sharma},
      year={2024},
      eprint={2308.12264},
      archivePrefix={arXiv},
      primaryClass={cs.LG},
      url={https://arxiv.org/abs/2308.12264}, 
}

@misc{anthony2020carbontrackertrackingpredictingcarbon,
      title={Carbontracker: Tracking and Predicting the Carbon Footprint of Training Deep Learning Models}, 
      author={Lasse F. Wolff Anthony and Benjamin Kanding and Raghavendra Selvan},
      year={2020},
      eprint={2007.03051},
      archivePrefix={arXiv},
      primaryClass={cs.CY},
      url={https://arxiv.org/abs/2007.03051}, 
}

@article{JMLR:v24:23-0069,
  author  = {Alexandra Sasha Luccioni and Sylvain Viguier and Anne-Laure Ligozat},
  title   = {Estimating the Carbon Footprint of BLOOM, a 176B Parameter Language Model},
  journal = {Journal of Machine Learning Research},
  year    = {2023},
  volume  = {24},
  number  = {253},
  pages   = {1--15},
  url     = {http://jmlr.org/papers/v24/23-0069.html}
}

@misc{jegham2025hungryaibenchmarkingenergy,
      title={How Hungry is AI? Benchmarking Energy, Water, and Carbon Footprint of LLM Inference}, 
      author={Nidhal Jegham and Marwan Abdelatti and Lassad Elmoubarki and Abdeltawab Hendawi},
      year={2025},
      eprint={2505.09598},
      archivePrefix={arXiv},
      primaryClass={cs.CY},
      url={https://arxiv.org/abs/2505.09598}, 
}

@article{greenAIstudy,
author = {Bol\'{o}n-Canedo, Ver\'{o}nica and Mor\'{a}n-Fern\'{a}ndez, Laura and Cancela, Brais and Alonso-Betanzos, Amparo},
title = {A review of green artificial intelligence: Towards a more sustainable future},
year = {2024},
issue_date = {Sep 2024},
publisher = {Elsevier Science Publishers B. V.},
address = {NLD},
volume = {599},
number = {C},
issn = {0925-2312},
url = {https://doi.org/10.1016/j.neucom.2024.128096},
doi = {10.1016/j.neucom.2024.128096},
journal = {Neurocomput.},
month = sep,
numpages = {10}
}

@Article{greenAI3,
AUTHOR = {Różycki, Rafał and Solarska, Dorota Agnieszka and Waligóra, Grzegorz},
TITLE = {Energy-Aware Machine Learning Models—A Review of Recent Techniques and Perspectives},
JOURNAL = {Energies},
VOLUME = {18},
YEAR = {2025},
NUMBER = {11},
ARTICLE-NUMBER = {2810},
URL = {https://www.mdpi.com/1996-1073/18/11/2810},
ISSN = {1996-1073},
DOI = {10.3390/en18112810}
}

@Article{greenAI2,
AUTHOR = {Sánchez-Mompó, Adrián and Mavromatis, Ioannis and Li, Peizheng and Katsaros, Konstantinos and Khan, Aftab},
TITLE = {Green MLOps to Green GenOps: An Empirical Study of Energy Consumption in Discriminative and Generative AI Operations},
JOURNAL = {Information},
VOLUME = {16},
YEAR = {2025},
NUMBER = {4},
ARTICLE-NUMBER = {281},
URL = {https://www.mdpi.com/2078-2489/16/4/281},
ISSN = {2078-2489},
DOI = {10.3390/info16040281}
}

@misc{fernandez2025energyconsiderationslargelanguage,
      title={Energy Considerations of Large Language Model Inference and Efficiency Optimizations}, 
      author={Jared Fernandez and Clara Na and Vashisth Tiwari and Yonatan Bisk and Sasha Luccioni and Emma Strubell},
      year={2025},
      eprint={2504.17674},
      archivePrefix={arXiv},
      primaryClass={cs.CL},
      url={https://arxiv.org/abs/2504.17674}, 
}

@misc{vaswani2023attentionneed,
      title={Attention Is All You Need}, 
      author={Ashish Vaswani and Noam Shazeer and Niki Parmar and Jakob Uszkoreit and Llion Jones and Aidan N. Gomez and Lukasz Kaiser and Illia Polosukhin},
      year={2023},
      eprint={1706.03762},
      archivePrefix={arXiv},
      primaryClass={cs.CL},
      url={https://arxiv.org/abs/1706.03762}, 
}

@misc{s2024cppjoulesenergymeasurementtool,
      title={CPPJoules: An Energy Measurement Tool for C++}, 
      author={Shivadharshan S and Akilesh P and Rajrupa Chattaraj and Sridhar Chimalakonda},
      year={2024},
      eprint={2412.13555},
      archivePrefix={arXiv},
      primaryClass={cs.SE},
      url={https://arxiv.org/abs/2412.13555}, 
}

@INPROCEEDINGS{10793163,
  author={Yang, Zeyu and Adamek, Karel and Armour, Wesley},
  booktitle={SC24: International Conference for High Performance Computing, Networking, Storage and Analysis}, 
  title={Accurate and Convenient Energy Measurements for GPUs: A Detailed Study of NVIDIA GPU’s Built-In Power Sensor}, 
  year={2024},
  volume={},
  number={},
  pages={1-17},
  doi={10.1109/SC41406.2024.00028}}

@misc{getzner2023accuracymetricmattersestimating,
      title={Accuracy is not the only Metric that matters: Estimating the Energy Consumption of Deep Learning Models}, 
      author={Johannes Getzner and Bertrand Charpentier and Stephan Günnemann},
      year={2023},
      eprint={2304.00897},
      archivePrefix={arXiv},
      primaryClass={cs.LG},
      url={https://arxiv.org/abs/2304.00897}, 
}

@article{luccioni2025aienergyscore,
  author       = {Sasha Luccioni and collaborators},
  title        = {Announcing {AI Energy Score Ratings}},
  year         = {2025},
  howpublished = {\url{https://huggingface.co/blog/sasha/announcing-ai-energy-score}},
  note         = {Accessed: 2025-09-06}
}

@misc{tian2025attentionsmicroscopecomparativestudy,
      title={Attentions Under the Microscope: A Comparative Study of Resource Utilization for Variants of Self-Attention}, 
      author={Zhengyu Tian and Anantha Padmanaban Krishna Kumar and Hemant Krishnakumar and Reza Rawassizadeh},
      year={2025},
      eprint={2507.07247},
      archivePrefix={arXiv},
      primaryClass={cs.LG},
      url={https://arxiv.org/abs/2507.07247}, 
}

@misc{pinnock2025edgeprofilerfastprofilingframework,
      title={EdgeProfiler: A Fast Profiling Framework for Lightweight LLMs on Edge Using Analytical Model}, 
      author={Alyssa Pinnock and Shakya Jayakody and Kawsher A Roxy and Md Rubel Ahmed},
      year={2025},
      eprint={2506.09061},
      archivePrefix={arXiv},
      primaryClass={cs.DC},
      url={https://arxiv.org/abs/2506.09061}, 
}

@misc{grattafiori2024llama3herdmodels,
      title={The Llama 3 Herd of Models}, 
      author={Aaron Grattafiori and Abhimanyu Dubey and Abhinav Jauhri and Abhinav Pandey and Abhishek Kadian and Ahmad Al-Dahle and Aiesha Letman and Akhil Mathur and Alan Schelten and Alex Vaughan and Amy Yang and Angela Fan and Anirudh Goyal and Anthony Hartshorn and Aobo Yang and Archi Mitra and Archie Sravankumar and Artem Korenev and Arthur Hinsvark and Arun Rao and Aston Zhang and Aurelien Rodriguez and Austen Gregerson and Ava Spataru and Baptiste Roziere and Bethany Biron and Binh Tang and Bobbie Chern and Charlotte Caucheteux and Chaya Nayak and Chloe Bi and Chris Marra and Chris McConnell and Christian Keller and Christophe Touret and Chunyang Wu and Corinne Wong and Cristian Canton Ferrer and Cyrus Nikolaidis and Damien Allonsius and Daniel Song and Danielle Pintz and Danny Livshits and Danny Wyatt and David Esiobu and Dhruv Choudhary and Dhruv Mahajan and Diego Garcia-Olano and Diego Perino and Dieuwke Hupkes and Egor Lakomkin and Ehab AlBadawy and Elina Lobanova and Emily Dinan and Eric Michael Smith and Filip Radenovic and Francisco Guzmán and Frank Zhang and Gabriel Synnaeve and Gabrielle Lee and Georgia Lewis Anderson and Govind Thattai and Graeme Nail and Gregoire Mialon and Guan Pang and Guillem Cucurell and Hailey Nguyen and Hannah Korevaar and Hu Xu and Hugo Touvron and Iliyan Zarov and Imanol Arrieta Ibarra and Isabel Kloumann and Ishan Misra and Ivan Evtimov and Jack Zhang and Jade Copet and Jaewon Lee and Jan Geffert and Jana Vranes and Jason Park and Jay Mahadeokar and Jeet Shah and Jelmer van der Linde and Jennifer Billock and Jenny Hong and Jenya Lee and Jeremy Fu and Jianfeng Chi and Jianyu Huang and Jiawen Liu and Jie Wang and Jiecao Yu and Joanna Bitton and Joe Spisak and Jongsoo Park and Joseph Rocca and Joshua Johnstun and Joshua Saxe and Junteng Jia and Kalyan Vasuden Alwala and Karthik Prasad and Kartikeya Upasani and Kate Plawiak and Ke Li and Kenneth Heafield and Kevin Stone and Khalid El-Arini and Krithika Iyer and Kshitiz Malik and Kuenley Chiu and Kunal Bhalla and Kushal Lakhotia and Lauren Rantala-Yeary and Laurens van der Maaten and Lawrence Chen and Liang Tan and Liz Jenkins and Louis Martin and Lovish Madaan and Lubo Malo and Lukas Blecher and Lukas Landzaat and Luke de Oliveira and Madeline Muzzi and Mahesh Pasupuleti and Mannat Singh and Manohar Paluri and Marcin Kardas and Maria Tsimpoukelli and Mathew Oldham and Mathieu Rita and Maya Pavlova and Melanie Kambadur and Mike Lewis and Min Si and Mitesh Kumar Singh and Mona Hassan and Naman Goyal and Narjes Torabi and Nikolay Bashlykov and Nikolay Bogoychev and Niladri Chatterji and Ning Zhang and Olivier Duchenne and Onur Çelebi and Patrick Alrassy and Pengchuan Zhang and Pengwei Li and Petar Vasic and Peter Weng and Prajjwal Bhargava and Pratik Dubal and Praveen Krishnan and Punit Singh Koura and Puxin Xu and Qing He and Qingxiao Dong and Ragavan Srinivasan and Raj Ganapathy and Ramon Calderer and Ricardo Silveira Cabral and Robert Stojnic and Roberta Raileanu and Rohan Maheswari and Rohit Girdhar and Rohit Patel and Romain Sauvestre and Ronnie Polidoro and Roshan Sumbaly and Ross Taylor and Ruan Silva and Rui Hou and Rui Wang and Saghar Hosseini and Sahana Chennabasappa and Sanjay Singh and Sean Bell and Seohyun Sonia Kim and Sergey Edunov and Shaoliang Nie and Sharan Narang and Sharath Raparthy and Sheng Shen and Shengye Wan and Shruti Bhosale and Shun Zhang and Simon Vandenhende and Soumya Batra and Spencer Whitman and Sten Sootla and Stephane Collot and Suchin Gururangan and Sydney Borodinsky and Tamar Herman and Tara Fowler and Tarek Sheasha and Thomas Georgiou and Thomas Scialom and Tobias Speckbacher and Todor Mihaylov and Tong Xiao and Ujjwal Karn and Vedanuj Goswami and Vibhor Gupta and Vignesh Ramanathan and Viktor Kerkez and Vincent Gonguet and Virginie Do and Vish Vogeti and Vítor Albiero and Vladan Petrovic and Weiwei Chu and Wenhan Xiong and Wenyin Fu and Whitney Meers and Xavier Martinet and Xiaodong Wang and Xiaofang Wang and Xiaoqing Ellen Tan and Xide Xia and Xinfeng Xie and Xuchao Jia and Xuewei Wang and Yaelle Goldschlag and Yashesh Gaur and Yasmine Babaei and Yi Wen and Yiwen Song and Yuchen Zhang and Yue Li and Yuning Mao and Zacharie Delpierre Coudert and Zheng Yan and Zhengxing Chen and Zoe Papakipos and Aaditya Singh and Aayushi Srivastava and Abha Jain and Adam Kelsey and Adam Shajnfeld and Adithya Gangidi and Adolfo Victoria and Ahuva Goldstand and Ajay Menon and Ajay Sharma and Alex Boesenberg and Alexei Baevski and Allie Feinstein and Amanda Kallet and Amit Sangani and Amos Teo and Anam Yunus and Andrei Lupu and Andres Alvarado and Andrew Caples and Andrew Gu and Andrew Ho and Andrew Poulton and Andrew Ryan and Ankit Ramchandani and Annie Dong and Annie Franco and Anuj Goyal and Aparajita Saraf and Arkabandhu Chowdhury and Ashley Gabriel and Ashwin Bharambe and Assaf Eisenman and Azadeh Yazdan and Beau James and Ben Maurer and Benjamin Leonhardi and Bernie Huang and Beth Loyd and Beto De Paola and Bhargavi Paranjape and Bing Liu and Bo Wu and Boyu Ni and Braden Hancock and Bram Wasti and Brandon Spence and Brani Stojkovic and Brian Gamido and Britt Montalvo and Carl Parker and Carly Burton and Catalina Mejia and Ce Liu and Changhan Wang and Changkyu Kim and Chao Zhou and Chester Hu and Ching-Hsiang Chu and Chris Cai and Chris Tindal and Christoph Feichtenhofer and Cynthia Gao and Damon Civin and Dana Beaty and Daniel Kreymer and Daniel Li and David Adkins and David Xu and Davide Testuggine and Delia David and Devi Parikh and Diana Liskovich and Didem Foss and Dingkang Wang and Duc Le and Dustin Holland and Edward Dowling and Eissa Jamil and Elaine Montgomery and Eleonora Presani and Emily Hahn and Emily Wood and Eric-Tuan Le and Erik Brinkman and Esteban Arcaute and Evan Dunbar and Evan Smothers and Fei Sun and Felix Kreuk and Feng Tian and Filippos Kokkinos and Firat Ozgenel and Francesco Caggioni and Frank Kanayet and Frank Seide and Gabriela Medina Florez and Gabriella Schwarz and Gada Badeer and Georgia Swee and Gil Halpern and Grant Herman and Grigory Sizov and Guangyi and Zhang and Guna Lakshminarayanan and Hakan Inan and Hamid Shojanazeri and Han Zou and Hannah Wang and Hanwen Zha and Haroun Habeeb and Harrison Rudolph and Helen Suk and Henry Aspegren and Hunter Goldman and Hongyuan Zhan and Ibrahim Damlaj and Igor Molybog and Igor Tufanov and Ilias Leontiadis and Irina-Elena Veliche and Itai Gat and Jake Weissman and James Geboski and James Kohli and Janice Lam and Japhet Asher and Jean-Baptiste Gaya and Jeff Marcus and Jeff Tang and Jennifer Chan and Jenny Zhen and Jeremy Reizenstein and Jeremy Teboul and Jessica Zhong and Jian Jin and Jingyi Yang and Joe Cummings and Jon Carvill and Jon Shepard and Jonathan McPhie and Jonathan Torres and Josh Ginsburg and Junjie Wang and Kai Wu and Kam Hou U and Karan Saxena and Kartikay Khandelwal and Katayoun Zand and Kathy Matosich and Kaushik Veeraraghavan and Kelly Michelena and Keqian Li and Kiran Jagadeesh and Kun Huang and Kunal Chawla and Kyle Huang and Lailin Chen and Lakshya Garg and Lavender A and Leandro Silva and Lee Bell and Lei Zhang and Liangpeng Guo and Licheng Yu and Liron Moshkovich and Luca Wehrstedt and Madian Khabsa and Manav Avalani and Manish Bhatt and Martynas Mankus and Matan Hasson and Matthew Lennie and Matthias Reso and Maxim Groshev and Maxim Naumov and Maya Lathi and Meghan Keneally and Miao Liu and Michael L. Seltzer and Michal Valko and Michelle Restrepo and Mihir Patel and Mik Vyatskov and Mikayel Samvelyan and Mike Clark and Mike Macey and Mike Wang and Miquel Jubert Hermoso and Mo Metanat and Mohammad Rastegari and Munish Bansal and Nandhini Santhanam and Natascha Parks and Natasha White and Navyata Bawa and Nayan Singhal and Nick Egebo and Nicolas Usunier and Nikhil Mehta and Nikolay Pavlovich Laptev and Ning Dong and Norman Cheng and Oleg Chernoguz and Olivia Hart and Omkar Salpekar and Ozlem Kalinli and Parkin Kent and Parth Parekh and Paul Saab and Pavan Balaji and Pedro Rittner and Philip Bontrager and Pierre Roux and Piotr Dollar and Polina Zvyagina and Prashant Ratanchandani and Pritish Yuvraj and Qian Liang and Rachad Alao and Rachel Rodriguez and Rafi Ayub and Raghotham Murthy and Raghu Nayani and Rahul Mitra and Rangaprabhu Parthasarathy and Raymond Li and Rebekkah Hogan and Robin Battey and Rocky Wang and Russ Howes and Ruty Rinott and Sachin Mehta and Sachin Siby and Sai Jayesh Bondu and Samyak Datta and Sara Chugh and Sara Hunt and Sargun Dhillon and Sasha Sidorov and Satadru Pan and Saurabh Mahajan and Saurabh Verma and Seiji Yamamoto and Sharadh Ramaswamy and Shaun Lindsay and Shaun Lindsay and Sheng Feng and Shenghao Lin and Shengxin Cindy Zha and Shishir Patil and Shiva Shankar and Shuqiang Zhang and Shuqiang Zhang and Sinong Wang and Sneha Agarwal and Soji Sajuyigbe and Soumith Chintala and Stephanie Max and Stephen Chen and Steve Kehoe and Steve Satterfield and Sudarshan Govindaprasad and Sumit Gupta and Summer Deng and Sungmin Cho and Sunny Virk and Suraj Subramanian and Sy Choudhury and Sydney Goldman and Tal Remez and Tamar Glaser and Tamara Best and Thilo Koehler and Thomas Robinson and Tianhe Li and Tianjun Zhang and Tim Matthews and Timothy Chou and Tzook Shaked and Varun Vontimitta and Victoria Ajayi and Victoria Montanez and Vijai Mohan and Vinay Satish Kumar and Vishal Mangla and Vlad Ionescu and Vlad Poenaru and Vlad Tiberiu Mihailescu and Vladimir Ivanov and Wei Li and Wenchen Wang and Wenwen Jiang and Wes Bouaziz and Will Constable and Xiaocheng Tang and Xiaojian Wu and Xiaolan Wang and Xilun Wu and Xinbo Gao and Yaniv Kleinman and Yanjun Chen and Ye Hu and Ye Jia and Ye Qi and Yenda Li and Yilin Zhang and Ying Zhang and Yossi Adi and Youngjin Nam and Yu and Wang and Yu Zhao and Yuchen Hao and Yundi Qian and Yunlu Li and Yuzi He and Zach Rait and Zachary DeVito and Zef Rosnbrick and Zhaoduo Wen and Zhenyu Yang and Zhiwei Zhao and Zhiyu Ma},
      year={2024},
      eprint={2407.21783},
      archivePrefix={arXiv},
      primaryClass={cs.AI},
      url={https://arxiv.org/abs/2407.21783}, 
}

@misc{openai2024gpt4technicalreport,
      title={GPT-4 Technical Report}, 
      author={OpenAI and Josh Achiam and Steven Adler and Sandhini Agarwal and Lama Ahmad and Ilge Akkaya and Florencia Leoni Aleman and Diogo Almeida and Janko Altenschmidt and Sam Altman and Shyamal Anadkat and Red Avila and Igor Babuschkin and Suchir Balaji and Valerie Balcom and Paul Baltescu and Haiming Bao and Mohammad Bavarian and Jeff Belgum and Irwan Bello and Jake Berdine and Gabriel Bernadett-Shapiro and Christopher Berner and Lenny Bogdonoff and Oleg Boiko and Madelaine Boyd and Anna-Luisa Brakman and Greg Brockman and Tim Brooks and Miles Brundage and Kevin Button and Trevor Cai and Rosie Campbell and Andrew Cann and Brittany Carey and Chelsea Carlson and Rory Carmichael and Brooke Chan and Che Chang and Fotis Chantzis and Derek Chen and Sully Chen and Ruby Chen and Jason Chen and Mark Chen and Ben Chess and Chester Cho and Casey Chu and Hyung Won Chung and Dave Cummings and Jeremiah Currier and Yunxing Dai and Cory Decareaux and Thomas Degry and Noah Deutsch and Damien Deville and Arka Dhar and David Dohan and Steve Dowling and Sheila Dunning and Adrien Ecoffet and Atty Eleti and Tyna Eloundou and David Farhi and Liam Fedus and Niko Felix and Simón Posada Fishman and Juston Forte and Isabella Fulford and Leo Gao and Elie Georges and Christian Gibson and Vik Goel and Tarun Gogineni and Gabriel Goh and Rapha Gontijo-Lopes and Jonathan Gordon and Morgan Grafstein and Scott Gray and Ryan Greene and Joshua Gross and Shixiang Shane Gu and Yufei Guo and Chris Hallacy and Jesse Han and Jeff Harris and Yuchen He and Mike Heaton and Johannes Heidecke and Chris Hesse and Alan Hickey and Wade Hickey and Peter Hoeschele and Brandon Houghton and Kenny Hsu and Shengli Hu and Xin Hu and Joost Huizinga and Shantanu Jain and Shawn Jain and Joanne Jang and Angela Jiang and Roger Jiang and Haozhun Jin and Denny Jin and Shino Jomoto and Billie Jonn and Heewoo Jun and Tomer Kaftan and Łukasz Kaiser and Ali Kamali and Ingmar Kanitscheider and Nitish Shirish Keskar and Tabarak Khan and Logan Kilpatrick and Jong Wook Kim and Christina Kim and Yongjik Kim and Jan Hendrik Kirchner and Jamie Kiros and Matt Knight and Daniel Kokotajlo and Łukasz Kondraciuk and Andrew Kondrich and Aris Konstantinidis and Kyle Kosic and Gretchen Krueger and Vishal Kuo and Michael Lampe and Ikai Lan and Teddy Lee and Jan Leike and Jade Leung and Daniel Levy and Chak Ming Li and Rachel Lim and Molly Lin and Stephanie Lin and Mateusz Litwin and Theresa Lopez and Ryan Lowe and Patricia Lue and Anna Makanju and Kim Malfacini and Sam Manning and Todor Markov and Yaniv Markovski and Bianca Martin and Katie Mayer and Andrew Mayne and Bob McGrew and Scott Mayer McKinney and Christine McLeavey and Paul McMillan and Jake McNeil and David Medina and Aalok Mehta and Jacob Menick and Luke Metz and Andrey Mishchenko and Pamela Mishkin and Vinnie Monaco and Evan Morikawa and Daniel Mossing and Tong Mu and Mira Murati and Oleg Murk and David Mély and Ashvin Nair and Reiichiro Nakano and Rajeev Nayak and Arvind Neelakantan and Richard Ngo and Hyeonwoo Noh and Long Ouyang and Cullen O'Keefe and Jakub Pachocki and Alex Paino and Joe Palermo and Ashley Pantuliano and Giambattista Parascandolo and Joel Parish and Emy Parparita and Alex Passos and Mikhail Pavlov and Andrew Peng and Adam Perelman and Filipe de Avila Belbute Peres and Michael Petrov and Henrique Ponde de Oliveira Pinto and Michael and Pokorny and Michelle Pokrass and Vitchyr H. Pong and Tolly Powell and Alethea Power and Boris Power and Elizabeth Proehl and Raul Puri and Alec Radford and Jack Rae and Aditya Ramesh and Cameron Raymond and Francis Real and Kendra Rimbach and Carl Ross and Bob Rotsted and Henri Roussez and Nick Ryder and Mario Saltarelli and Ted Sanders and Shibani Santurkar and Girish Sastry and Heather Schmidt and David Schnurr and John Schulman and Daniel Selsam and Kyla Sheppard and Toki Sherbakov and Jessica Shieh and Sarah Shoker and Pranav Shyam and Szymon Sidor and Eric Sigler and Maddie Simens and Jordan Sitkin and Katarina Slama and Ian Sohl and Benjamin Sokolowsky and Yang Song and Natalie Staudacher and Felipe Petroski Such and Natalie Summers and Ilya Sutskever and Jie Tang and Nikolas Tezak and Madeleine B. Thompson and Phil Tillet and Amin Tootoonchian and Elizabeth Tseng and Preston Tuggle and Nick Turley and Jerry Tworek and Juan Felipe Cerón Uribe and Andrea Vallone and Arun Vijayvergiya and Chelsea Voss and Carroll Wainwright and Justin Jay Wang and Alvin Wang and Ben Wang and Jonathan Ward and Jason Wei and CJ Weinmann and Akila Welihinda and Peter Welinder and Jiayi Weng and Lilian Weng and Matt Wiethoff and Dave Willner and Clemens Winter and Samuel Wolrich and Hannah Wong and Lauren Workman and Sherwin Wu and Jeff Wu and Michael Wu and Kai Xiao and Tao Xu and Sarah Yoo and Kevin Yu and Qiming Yuan and Wojciech Zaremba and Rowan Zellers and Chong Zhang and Marvin Zhang and Shengjia Zhao and Tianhao Zheng and Juntang Zhuang and William Zhuk and Barret Zoph},
      year={2024},
      eprint={2303.08774},
      archivePrefix={arXiv},
      primaryClass={cs.CL},
      url={https://arxiv.org/abs/2303.08774}, 
}

@misc{yang2025qwen3technicalreport,
      title={Qwen3 Technical Report}, 
      author={An Yang and Anfeng Li and Baosong Yang and Beichen Zhang and Binyuan Hui and Bo Zheng and Bowen Yu and Chang Gao and Chengen Huang and Chenxu Lv and Chujie Zheng and Dayiheng Liu and Fan Zhou and Fei Huang and Feng Hu and Hao Ge and Haoran Wei and Huan Lin and Jialong Tang and Jian Yang and Jianhong Tu and Jianwei Zhang and Jianxin Yang and Jiaxi Yang and Jing Zhou and Jingren Zhou and Junyang Lin and Kai Dang and Keqin Bao and Kexin Yang and Le Yu and Lianghao Deng and Mei Li and Mingfeng Xue and Mingze Li and Pei Zhang and Peng Wang and Qin Zhu and Rui Men and Ruize Gao and Shixuan Liu and Shuang Luo and Tianhao Li and Tianyi Tang and Wenbiao Yin and Xingzhang Ren and Xinyu Wang and Xinyu Zhang and Xuancheng Ren and Yang Fan and Yang Su and Yichang Zhang and Yinger Zhang and Yu Wan and Yuqiong Liu and Zekun Wang and Zeyu Cui and Zhenru Zhang and Zhipeng Zhou and Zihan Qiu},
      year={2025},
      eprint={2505.09388},
      archivePrefix={arXiv},
      primaryClass={cs.CL},
      url={https://arxiv.org/abs/2505.09388}, 
}

@misc{cao2021ireneinterpretableenergyprediction,
      title={IrEne: Interpretable Energy Prediction for Transformers}, 
      author={Qingqing Cao and Yash Kumar Lal and Harsh Trivedi and Aruna Balasubramanian and Niranjan Balasubramanian},
      year={2021},
      eprint={2106.01199},
      archivePrefix={arXiv},
      primaryClass={cs.CL},
      url={https://arxiv.org/abs/2106.01199}, 
}

@misc{schwartz2019greenai,
      title={Green AI}, 
      author={Roy Schwartz and Jesse Dodge and Noah A. Smith and Oren Etzioni},
      year={2019},
      eprint={1907.10597},
      archivePrefix={arXiv},
      primaryClass={cs.CY},
      url={https://arxiv.org/abs/1907.10597}, 
}

@misc{vandervlugt2025powersensor3fastaccurateopen,
      title={PowerSensor3: A Fast and Accurate Open Source Power Measurement Tool}, 
      author={Steven van der Vlugt and Leon Oostrum and Gijs Schoonderbeek and Ben van Werkhoven and Bram Veenboer and Krijn Doekemeijer and John W. Romein},
      year={2025},
      eprint={2504.17883},
      archivePrefix={arXiv},
      primaryClass={cs.PF},
      url={https://arxiv.org/abs/2504.17883}, 
}

@article{ALVI2021100594,
title = {MLEE: Method Level Energy Estimation — A machine learning approach},
journal = {Sustainable Computing: Informatics and Systems},
volume = {32},
pages = {100594},
year = {2021},
issn = {2210-5379},
doi = {https://doi.org/10.1016/j.suscom.2021.100594},
url = {https://www.sciencedirect.com/science/article/pii/S2210537921000822},
author = {Hamza Mustafa Alvi and Hammad Majeed and Hasan Mujtaba and Mirza Omer Beg}
}

@inproceedings{Casta_o_2023,
   title={Exploring the Carbon Footprint of Hugging Face’s ML Models: A Repository Mining Study},
   url={http://dx.doi.org/10.1109/ESEM56168.2023.10304801},
   DOI={10.1109/esem56168.2023.10304801},
   booktitle={2023 ACM/IEEE International Symposium on Empirical Software Engineering and Measurement (ESEM)},
   publisher={IEEE},
   author={Castaño, Joel and Martínez-Fernández, Silverio and Franch, Xavier and Bogner, Justus},
   year={2023},
   month=oct, pages={1–12} }

@article{10.1145/2962131,
author = {Bridges, Robert A. and Imam, Neena and Mintz, Tiffany M.},
title = {Understanding GPU Power: A Survey of Profiling, Modeling, and Simulation Methods},
year = {2016},
issue_date = {September 2017},
publisher = {Association for Computing Machinery},
address = {New York, NY, USA},
volume = {49},
number = {3},
issn = {0360-0300},
url = {https://doi.org/10.1145/2962131},
doi = {10.1145/2962131},
journal = {ACM Comput. Surv.},
month = sep,
articleno = {41},
numpages = {27}
}

@misc{ozcan2025quantifyingenergyconsumptioncarbon,
      title={Quantifying the Energy Consumption and Carbon Emissions of LLM Inference via Simulations}, 
      author={Miray Özcan and Philipp Wiesner and Philipp Weiß and Odej Kao},
      year={2025},
      eprint={2507.11417},
      archivePrefix={arXiv},
      primaryClass={cs.DC},
      url={https://arxiv.org/abs/2507.11417}, 
}

@misc{faiz2024llmcarbonmodelingendtoendcarbon,
      title={LLMCarbon: Modeling the end-to-end Carbon Footprint of Large Language Models}, 
      author={Ahmad Faiz and Sotaro Kaneda and Ruhan Wang and Rita Osi and Prateek Sharma and Fan Chen and Lei Jiang},
      year={2024},
      eprint={2309.14393},
      archivePrefix={arXiv},
      primaryClass={cs.CL},
      url={https://arxiv.org/abs/2309.14393}, 
}

@inproceedings{Arafa_2020, series={CF ’20},
   title={Verified instruction-level energy consumption measurement for NVIDIA GPUs},
   url={http://dx.doi.org/10.1145/3387902.3392613},
   DOI={10.1145/3387902.3392613},
   booktitle={Proceedings of the 17th ACM International Conference on Computing Frontiers},
   publisher={ACM},
   author={Arafa, Yehia and ElWazir, Ammar and ElKanishy, Abdelrahman and Aly, Youssef and Elsayed, Ayatelrahman and Badawy, Abdel-Hameed and Chennupati, Gopinath and Eidenbenz, Stephan and Santhi, Nandakishore},
   year={2020},
   month=may, pages={60–70},
   collection={CF ’20} }

@article{article,
author = {Stone, Peter and Veloso, Manuela},
year = {2000},
month = {05},
pages = {},
title = {Multiagent Systems: A Survey from a Machine Learning Perspective},
volume = {8},
journal = {Autonomous Robots},
doi = {10.1023/A:1008942012299}
}

@misc{PNY-RTX5000AdaDatasheet,
  author       = {PNY},
  title        = {NVIDIA RTX 5000 Ada Generation Datasheet},
  year         = {2023},
  howpublished = {Product Datasheet (PDF)},
  url          = {https://www.pny.com/en-eu/File%20Library/Professional/DATASHEET/WORKSTATION/PNY-NVIDIA-RTX-5000-Ada-Generation-Datasheet.pdf},
  note         = {Specifications include CUDA cores, Tensor/RT cores, memory capacity, bandwidth, and board power.}
}

@misc{NVIDIA-RTX6000AdaDatasheet,
  author       = {NVIDIA},
  title        = {NVIDIA RTX 6000 Ada Generation Datasheet},
  year         = {2023},
  howpublished = {Technical Datasheet (PDF)},
  url          = {https://www.nvidia.com/content/dam/en-zz/Solutions/design-visualization/rtx-6000/proviz-print-rtx6000-datasheet-web-2504660.pdf},
  note         = {Official specifications for the RTX 6000 Ada, including CUDA/RT/Tensor cores, memory, bandwidth, and power.}
}

@inproceedings{Yang_2024,
   title={Accurate and Convenient Energy Measurements for GPUs: A Detailed Study of NVIDIA GPU’s Built-In Power Sensor},
   url={http://dx.doi.org/10.1109/SC41406.2024.00028},
   DOI={10.1109/sc41406.2024.00028},
   booktitle={SC24: International Conference for High Performance Computing, Networking, Storage and Analysis},
   publisher={IEEE},
   author={Yang, Zeyu and Adamek, Karel and Armour, Wesley},
   year={2024},
   month=nov, pages={1–17} }

@misc{nik2025energyconsciousllmdecodingimpact,
      title={Energy-Conscious LLM Decoding: Impact of Text Generation Strategies on GPU Energy Consumption}, 
      author={Alireza Nik and Michael A. Riegler and Pål Halvorsen},
      year={2025},
      eprint={2502.11723},
      archivePrefix={arXiv},
      primaryClass={cs.AI},
      url={https://arxiv.org/abs/2502.11723}, 
}

@misc{bert,
      title={BERT: Pre-training of Deep Bidirectional Transformers for Language Understanding}, 
      author={Jacob Devlin and Ming-Wei Chang and Kenton Lee and Kristina Toutanova},
      year={2019},
      eprint={1810.04805},
      archivePrefix={arXiv},
      primaryClass={cs.CL},
      url={https://arxiv.org/abs/1810.04805}, 
}

@misc{albert,
      title={ALBERT: A Lite BERT for Self-supervised Learning of Language Representations}, 
      author={Zhenzhong Lan and Mingda Chen and Sebastian Goodman and Kevin Gimpel and Piyush Sharma and Radu Soricut},
      year={2020},
      eprint={1909.11942},
      archivePrefix={arXiv},
      primaryClass={cs.CL},
      url={https://arxiv.org/abs/1909.11942}, 
}

@misc{distilbert,
      title={DistilBERT, a distilled version of BERT: smaller, faster, cheaper and lighter}, 
      author={Victor Sanh and Lysandre Debut and Julien Chaumond and Thomas Wolf},
      year={2020},
      eprint={1910.01108},
      archivePrefix={arXiv},
      primaryClass={cs.CL},
      url={https://arxiv.org/abs/1910.01108}, 
}

@misc{roberta,
      title={RoBERTa: A Robustly Optimized BERT Pretraining Approach}, 
      author={Yinhan Liu and Myle Ott and Naman Goyal and Jingfei Du and Mandar Joshi and Danqi Chen and Omer Levy and Mike Lewis and Luke Zettlemoyer and Veselin Stoyanov},
      year={2019},
      eprint={1907.11692},
      archivePrefix={arXiv},
      primaryClass={cs.CL},
      url={https://arxiv.org/abs/1907.11692}, 
}

@article{llama3modelcard,

title={Llama 3 Model Card},

author={AI@Meta},

year={2024},

url = {https://github.com/meta-llama/llama3/blob/main/MODEL_CARD.md}

}

@misc{gemma,
      title={Gemma: Open Models Based on Gemini Research and Technology}, 
      author={Gemma Team},
      year={2024},
      eprint={2403.08295},
      archivePrefix={arXiv},
      primaryClass={cs.CL},
      url={https://arxiv.org/abs/2403.08295}, 
}

@misc{qwen2,
      title={Qwen2 Technical Report}, 
      author={An Yang and Baosong Yang and Binyuan Hui and Bo Zheng and Bowen Yu and Chang Zhou and Chengpeng Li and Chengyuan Li and Dayiheng Liu and Fei Huang and Guanting Dong and Haoran Wei and Huan Lin and Jialong Tang and Jialin Wang and Jian Yang and Jianhong Tu and Jianwei Zhang and Jianxin Ma and Jianxin Yang and Jin Xu and Jingren Zhou and Jinze Bai and Jinzheng He and Junyang Lin and Kai Dang and Keming Lu and Keqin Chen and Kexin Yang and Mei Li and Mingfeng Xue and Na Ni and Pei Zhang and Peng Wang and Ru Peng and Rui Men and Ruize Gao and Runji Lin and Shijie Wang and Shuai Bai and Sinan Tan and Tianhang Zhu and Tianhao Li and Tianyu Liu and Wenbin Ge and Xiaodong Deng and Xiaohuan Zhou and Xingzhang Ren and Xinyu Zhang and Xipin Wei and Xuancheng Ren and Xuejing Liu and Yang Fan and Yang Yao and Yichang Zhang and Yu Wan and Yunfei Chu and Yuqiong Liu and Zeyu Cui and Zhenru Zhang and Zhifang Guo and Zhihao Fan},
      year={2024},
      eprint={2407.10671},
      archivePrefix={arXiv},
      primaryClass={cs.CL},
      url={https://arxiv.org/abs/2407.10671}, 
}

@misc{abdin2024phi4technicalreport,
      title={Phi-4 Technical Report}, 
      author={Marah Abdin and Jyoti Aneja and Harkirat Behl and Sébastien Bubeck and Ronen Eldan and Suriya Gunasekar and Michael Harrison and Russell J. Hewett and Mojan Javaheripi and Piero Kauffmann and James R. Lee and Yin Tat Lee and Yuanzhi Li and Weishung Liu and Caio C. T. Mendes and Anh Nguyen and Eric Price and Gustavo de Rosa and Olli Saarikivi and Adil Salim and Shital Shah and Xin Wang and Rachel Ward and Yue Wu and Dingli Yu and Cyril Zhang and Yi Zhang},
      year={2024},
      eprint={2412.08905},
      archivePrefix={arXiv},
      primaryClass={cs.CL},
      url={https://arxiv.org/abs/2412.08905}, 
}

@article{bart,
  author       = {Mike Lewis and
                  Yinhan Liu and
                  Naman Goyal and
                  Marjan Ghazvininejad and
                  Abdelrahman Mohamed and
                  Omer Levy and
                  Veselin Stoyanov and
                  Luke Zettlemoyer},
  title        = {{BART:} Denoising Sequence-to-Sequence Pre-training for Natural Language
                  Generation, Translation, and Comprehension},
  journal      = {CoRR},
  volume       = {abs/1910.13461},
  year         = {2019},
  url          = {http://arxiv.org/abs/1910.13461},
  eprinttype    = {arXiv},
  eprint       = {1910.13461},
  bibsource    = {dblp computer science bibliography, https://dblp.org}
}

@misc{flant5,
      title={Scaling Instruction-Finetuned Language Models}, 
      author={Hyung Won Chung and Le Hou and Shayne Longpre and Barret Zoph and Yi Tay and William Fedus and Yunxuan Li and Xuezhi Wang and Mostafa Dehghani and Siddhartha Brahma and Albert Webson and Shixiang Shane Gu and Zhuyun Dai and Mirac Suzgun and Xinyun Chen and Aakanksha Chowdhery and Alex Castro-Ros and Marie Pellat and Kevin Robinson and Dasha Valter and Sharan Narang and Gaurav Mishra and Adams Yu and Vincent Zhao and Yanping Huang and Andrew Dai and Hongkun Yu and Slav Petrov and Ed H. Chi and Jeff Dean and Jacob Devlin and Adam Roberts and Denny Zhou and Quoc V. Le and Jason Wei},
      year={2022},
      eprint={2210.11416},
      archivePrefix={arXiv},
      primaryClass={cs.LG},
      url={https://arxiv.org/abs/2210.11416}, 
}

@misc{gptoss,
      title={gpt-oss-120b \& gpt-oss-20b Model Card}, 
      author={OpenAI},
      year={2025},
      eprint={2508.10925},
      archivePrefix={arXiv},
      primaryClass={cs.CL},
      url={https://arxiv.org/abs/2508.10925}, 
}
